\documentclass{article}

\PassOptionsToPackage{numbers, compress}{natbib}

\usepackage[final]{neurips_2024}

\usepackage[toc,page,header]{appendix}
\usepackage{natbib}
\usepackage{multirow}
\usepackage{booktabs}
\usepackage{graphicx}
\usepackage[table]{xcolor}
\usepackage{tabularray}
\usepackage{amsmath}
\usepackage{diagbox}
\usepackage{tabularx}
\usepackage{caption}
\usepackage{subcaption}
\usepackage{amssymb}
\usepackage{wrapfig}
\usepackage{minitoc}
\usepackage{fancyhdr}
\usepackage[utf8]{inputenc} 
\usepackage[T1]{fontenc}    
\usepackage{hyperref}       
\usepackage{url}            
\usepackage{amsfonts}       
\usepackage{nicefrac}       
\usepackage{microtype}      
\usepackage{udga}

\title{Unified Domain Generalization and Adaptation for Multi-View 3D Object Detection}

\author{
  Gyusam Chang$^1$$^*$ 
  \qquad Jiwon Lee$^2$\thanks{These authors contributed equally.} 
  \qquad Donghyun Kim$^1$ 
  \qquad Jinkyu Kim$^1$ \\
  \textbf{Dongwook Lee}$^2$ 
  \qquad \textbf{Daehyun Ji}$^2$ 
  \qquad \textbf{Sujin Jang}$^2$$^\dagger$ 
  \qquad \textbf{Sangpil Kim}$^1$\thanks{Corresponding authors.} \\
  $^1$Korea University \\
  $^2$Samsung Advanced Institute of Technology \\
  \texttt{$\{$gsjang95, d$\_$kim, jinkyukim, spk7$\}$@korea.ac.kr}\\
  \texttt{$\{$ji1.lee, dw12.lee, derek.ji, s.steve.jang$\}$@samsung.com}
  }

\begin{document}

\doparttoc 
\faketableofcontents 

\maketitle

\begin{abstract}

Recent advances in 3D object detection leveraging multi-view cameras have demonstrated their practical and economical value in various challenging vision tasks.
However, typical supervised learning approaches face challenges in achieving satisfactory adaptation toward unseen and unlabeled target datasets (\ie, direct transfer) due to the inevitable geometric misalignment between the source and target domains.
In practice, we also encounter constraints on resources for training models and collecting annotations for the successful deployment of 3D object detectors.
In this paper, we propose Unified Domain Generalization and Adaptation (UDGA), a practical solution to mitigate those drawbacks.
We first propose Multi-view Overlap Depth Constraint that leverages the strong association between multi-view, significantly alleviating geometric gaps due to perspective view changes.
Then, we present a Label-Efficient Domain Adaptation approach to handle unfamiliar targets with significantly fewer amounts of labels (\ie, 1$\%$ and 5$\%)$, while preserving well-defined source knowledge for training efficiency.
Overall, UDGA framework enables stable detection performance in both source and target domains, effectively bridging inevitable domain gaps, while demanding fewer annotations.
We demonstrate the robustness of UDGA with large-scale benchmarks: nuScenes, Lyft, and Waymo, where our framework outperforms the current state-of-the-art methods. 

\end{abstract}
\section{Introduction}
\label{introduction}

3D Object Detection (3DOD) is a pivotal computer vision task in various real-world applications such as autonomous driving and robotics. 
Recent progress in 3DOD~\cite{bai2022transfusion, chang2024cmda, liang2022bevfusion, liu2023bevfusion} have showcased remarkable advancements, primarily due to the large-scale benchmark datasets~\cite{sun2020scalability, caesar2020nuscenes, houston2021one} and the introduction of multiple computer vision sensors~(\eg, LiDAR, multi-view cameras, and RADAR).
Among these, camera-based multi-view 3DOD~\cite{wang2022detr3d, li2022bevformer, roh2022ora3d, huang2021bevdet, li2023bevdepth} has drawn significant attention for its cost-efficiency and rich semantic information.
However, a significant challenge remains largely unexplored: accurately detecting the location and category of objects in the presence of distributional shifts between the source and target domains (\ie, data distributional gaps between the training and the testing datasets).

\begin{figure}
    \centering
    \includegraphics[width=1.0\linewidth]{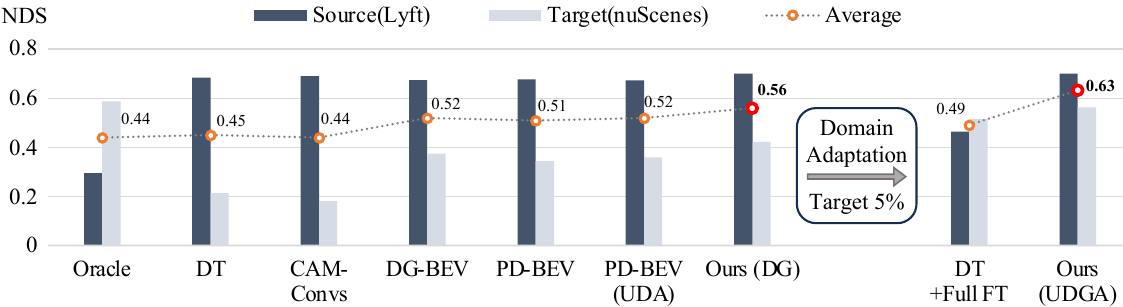}
    \caption{Comparison of performance in both source and target domains (Tab.~\ref{tab:UDGA2}). 
    Here, ``Average'' (orange dots) refers to mean NDS in both the source and target domains. 
    We draw comparisons with prior methods CAM-Conv~\cite{facil2019cam}, DG-BEV~\cite{wang2023towards} and PD-BEV~\cite{lu2023towards} offering an empirical lower and upper bounds, DT and Oracle.
    Note that we only use 5$\%$ of the target label for Domain Adaptation.}
    \label{fig:gda_graph}
\end{figure}

To successfully develop and deploy Multi-view 3DOD models, we need to solve two practical problems:
(1) the geometric distributional shift across different sensor configurations, and (2) the limited amount of resources (\eg, insufficient computing resources, expensive data annotations).
The first problem poses a challenge in learning transferable knowledge for robust generalization in novel domains.
The second issue inevitably requires efficient utilization of computing resources for training and inference, as well as label-efficient development of 3DOD models in practice.
To tackle these practical problems, we introduce a \textbf{U}nified \textbf{D}omain \textbf{G}eneralization and \textbf{A}daptation (UDGA) strategy, which addresses a series of domain shift problems (\ie, learning domain generalizable features significantly improves the quality of parameter- and label-efficient few-shot domain adaptation).

Prior studies aim to learn domain-agnostic knowledge alleviating domain shifts from drastic view changes in cross-domain environments.
DG-BEV~\cite{wang2023towards} disentangles the camera intrinsic parameters and trains the network with a domain discriminator for view-invariant feature learning.
Similarly, PD-BEV~\cite{lu2023towards} renders implicit foreground volumes and suppresses the perspective bias leveraging semantic supervision.
However, these approaches struggle to capture optimal representations, highlighting that there is still room for improvements in novel target domains (\ie, up to -50.8$\%$ Closed Gap compared to Oracle).
To tackle these drawbacks, we first advocate a Multi-view Overlap Depth Constraint that leverages occluded regions between adjacent views, which serve as notable triangular clues to guarantee geometric consistency.
This approach effectively addresses perspective differences between cross-domain environments by directly penalising the corresponding depth between adjacent views, and shows considerable generalization capacity (up to +75.8$\%$ Closed Gap compared to DT).

Nevertheless, the development of algorithms running on edge devices (\ie, autonomous vehicles) faces the challenge of limited resources, which requires efficient utilization of computing systems.
To resolve these challenges, we carefully design a \textit{go-to} strategy, Label-Efficient Domain Adaptation, that bridges two different domains with cost-effective transfer learning.
Precisely, motivated by Parameter-Efficient Fine-Tuning (PEFT)~\cite{hu2021lora, lester2021power, liu2021p}, we focus on smooth adaptation to target domains by fully exploiting well-defined source knowledge.
Specifically, leveraging plug-and-play extra parameters, we substantially adapt to target domains while retaining information from the source domain (+14$\%$ Average gain compared to DT+Full FT as shown in Fig.~\ref{fig:gda_graph}).
As a result, we note that UDGA practically expand base models, efficiently boosting overall capacity under limited resources.

Given landmark datasets in 3DOD, nuScenes~\cite{caesar2020nuscenes}, Lyft~\cite{houston2021one} and Waymo~\cite{sun2020scalability}, we validate the effectiveness of our UDGA framework for the camera-based multi-view 3DOD task.
Notably, we achieve state-of-the-art performance in cross-domain environments and demonstrate the component-wise effectiveness through ablation studies.
To summarize, our main contributions are as follows:
\begin{itemize}

\item We introduce the Unified Domain Generalization and Adaptation (UDGA) framework, which aims to learn generalizable geometric features and improve resource efficiency for enhanced practicality in addressing distributional shift alignments.

\item We advocate depth-scale consistency across multi-view images to effectively address 3D geometric misalignment problems. To this end, we leverage the corresponding triangular cues between adjacent views to seamlessly bridge the domain gap.

\item We present a label- and parameter-efficient domain adaptation method, which requires fewer annotations and fine-tuning parameters while preserving source-domain knowledge.

\item We demonstrate the effectiveness of UDGA on multiple challenging cross-domain benchmarks~(\ie, Lyft $\rightarrow$ nuScenes, nuScenes $\rightarrow$ Lyft, and Waymo $\rightarrow$ nuScenes). 
The results show that UDGA achieves a new state-of-the-art performance in Multi-view 3DOD.

\end{itemize}

\section{Related Work}
\label{related_works}

\subsection{Multi-view 3D Object Detection}
3D object detection~\cite{liu2023bevfusion, yan2018second, bai2022transfusion, park2021dd3d, wang2021pgd, wang2021fcos3d, brazil2019m3d,liu2020smoke, zhou2018voxelnet, lang2019pointpillars} is a fundamental aspect of computer vision tasks in the real world.
Especially, Multi-view 3D Object Detection leveraging Bird's Eye View (BEV) representations~\cite{huang2021bevdet, li2023bevdepth, wang2022detr3d} have rapidly expanded. 
We observe that this paradigm is divided into two categories: (i) LSS-based~\cite{philion2020lift, huang2021bevdet, li2023bevdepth}, and (ii) Query-based~\cite{wang2022detr3d, liu2022petr, roh2022ora3d}.
The former adopts explicit methods leveraging depth estimation network, and the latter concentrates on implicit methods utilizing the attention mechanism of Transformer~\cite{vaswani2017attention}.
Recently, these methods~\cite{li2022bevformer, park2022time, yang2023bevformer} significantly benefit from improved geometric understanding leveraging temporal inputs.
Also, methods~\cite{yuhong-CMKD-ECCV2022, huang2022tig, chen2022bevdistill, jang2024stxd} that directly guide the model using the LiDAR teacher model significantly encourage BEV spatial details.
In particular, this approach is being adopted to gradually replace LiDAR in real-world scenarios; however, it still suffers from poor generalizability due to drastic domain shifts (\eg, weather, country, and sensor). To mitigate these issues, we present a novel paradigm, Unsupervised Domain Generalization and Adaptation (UDGA), that effectively addresses geometric issues leveraging multi-view triangular clues and smoothly bridge differenet domains without forgetting previously learned knowledge.


\subsection{Bridging the Domain Gap for 3D Object Detection}
Due to the expensive cost of sophisticated sensor configurations and accurate 3D annotations for autonomous driving scenes, existing works strive to generalize 3D perception models in various data distributions. Specifically, they often fail to address the covariate shift between the training and test splits. To bridge the domain gap, existing approaches have introduced noteworthy solutions as below. 

\myparagraph{LiDAR-based.} Wang~\etal~\cite{wang2020train} introduced Statistical Normalization (SN) to mitigate the differences in object size distribution across various datasets. ST3D~\cite{yang2021st3d} leveraged domain knowledge through random object scale augmentation, and their self-training pipeline refined the pseudo-labels. SPG~\cite{xu2021spg} aims to capture the spatial shape, generating the missing points. 3D-CoCo~\cite{yihan2021learning} contrastively adjust the domain boundary between source and target to extract robust features. LiDAR Distillation ~\cite{wei2022lidar} generates pseudo sparse point sets in spherical coordinates and aligns the knowledge between source and pseudo target. STAL3D~\cite{zhang2024stal3d} effectively extended ST3D by incorporating adversarial learning. DTS~\cite{hu2023density} randomly re-sample the beam and aim to capture the cross-density between student and teacher models. CMDA~\cite{chang2024cmda} aim to learn rich-semantic knowledge from camera BEV features and adversarially guide seen sources and unseen targets, achieving state-of-the-art UDA capacity. 

\myparagraph{Camera-based.} While various groundbreaking methods based on LiDAR have been researched, camera-based approaches are still limited. Due to the elaborate 2D-3D alignment, not only are LiDAR-based approaches not directly applicable, but conventional 2D visual approaches~\cite{vidit2023clip, wu2022single, he2022masked, chen2020simple} cannot be adopted either. To mitigate these issues, STMono3D~\cite{li2022unsupervised} self-supervise the monocular 3D detection network in a teacher-student manner.
DG-BEV~\cite{wang2023towards} adversarially guide the network from perspective augmented multi-view images. 
PD-BEV~\cite{lu2023towards} explicitly supervise models by the RenderNet with pseudo labels. 
However, camera domain generalization methods cannot meet the performance required for the safety, struggling to address the practical domain shift in the perspective change. To narrow the gap, we introduce a Unified Domain Generalization and Adaptation (UDGA) framework that effectively promotes depth-scale consistency by leveraging occluded clues between adjacent views and then seamlessly transfers the model's potential along with a few novel labels.


\subsection{Parameter Efficient Fine-Tuning}
Recent NLP works fully benefit from general-purpose Large-language Models (LLM). 
Additionally, they have proposed Parameter-Efficient Fine-Tuning (PEFT)~\cite{lester2021power, hu2021lora, houlsby2019parameter, lian2022scaling, liu2022few} to effectively transfer LLM power to various downstream tasks.
Specifically, PEFT preserves and exploits previously learned universal information, fine-tuning only additional parameters with a few downstream labels.
This paradigm enables to notably reduce extensive computational resources, and large amounts of task-specific data and also effectively address challenging domain shifts in various downstream tasks as reported by~\cite{dong2024ppea}.
Inspired by this motivation, to address drastic perspective shifts between source and target domains, we design Label-Efficient Domain Adaptation that fully transfers generalized source potentials to target domains by fine-tuning only our extra modules with few-shot target data.

\section{Methodology}
\label{methodology}

\begin{figure}[t]
    \centering
    \includegraphics[width=1.0\linewidth]{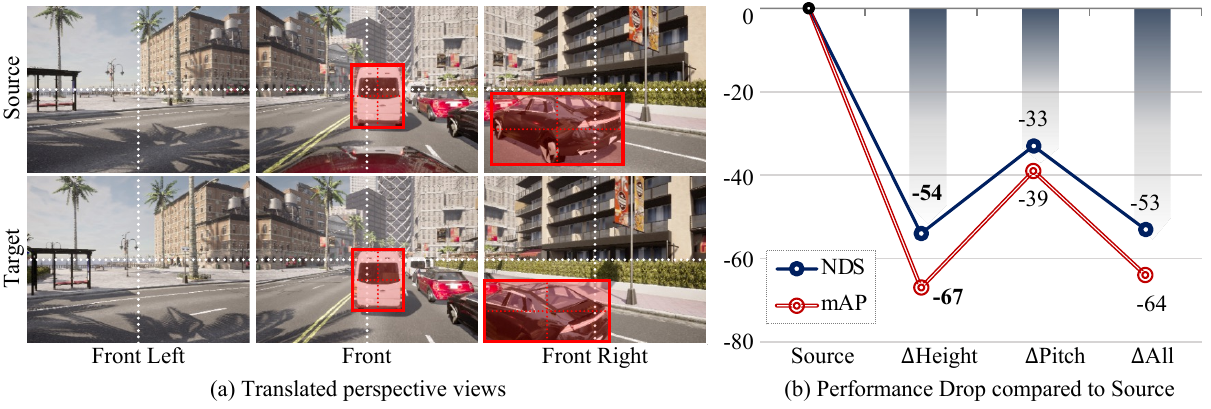}
    \caption{(a) An illustration of multi-view installation translation difference. The first (\ie, source) and second (\ie, target) rows are two perspective views of the same scene captured from different installation points. The translation gap between these views is substantial, approximately 30$\%$. (b) Source trained network shows poor perception capability in target domain, primarily due to extrinsic shifts. In $\Delta$Height, mAP and NDS have dropped up to -67$\%$ compared to source. Note that we simulate the camera extrinsic shift leveraging CARLA~\cite{Dosovitskiy17} (refer to Appendix~\ref{apx:dataset} for further details).}
    \label{fig:domain_shift}
\end{figure}


\subsection{Preliminary}
\label{3.1}
Multi-view 3D Object Detection is a fundamental computer vision task that involves safely localizing and categorizing objects in a 3D space exploiting 2D visual information from multiple camera views.
Especially, recent landmark Multi-view 3D Object Detection models~\cite{wang2022detr3d, roh2022ora3d, li2022bevformer, huang2021bevdet, huang2022tig} are formulated as follow; $\arg\min\mathcal{L}(Y, \mathcal{D}({\mathcal{V}(I,K,T)})$, where $Y$ represents the size $(l, w, h)$, centerness $(cx, cy, cz)$, and rotation $\phi$ of each 3D object. Also, $I=\{i_1,i_2,...,i_n\} \in \mathbb{R}^{N \times H \times W \times 3}$, $K$, and $T=[R|t]$ denotes multi-view images, intrinsic and extrinsic parameters. 
Specifically, these models, which fully benefit from view transformation modules $\mathcal{V}$, encode 2D visual features alongside the 3D spatial environment into a bird's eye view (BEV) representation. 
First, these works adopts explicit methods (BEV view transformation $\mathcal{P}$ as shown in Eq.~\ref{eq:lss}) exploiting depth estimation network.
Subsequently, Detector Head modules $\mathcal{D}$ supervises BEV features with 3D labels $Y$ in a three-dimensional manner. 
\begin{equation}
\label{eq:lss}
    \mathcal{V}(I,K,T) = \mathcal{P}(F_{2d} \otimes D, K, T),
\end{equation}

\begin{figure}[t!]
    \centering
    \includegraphics[width=\linewidth]{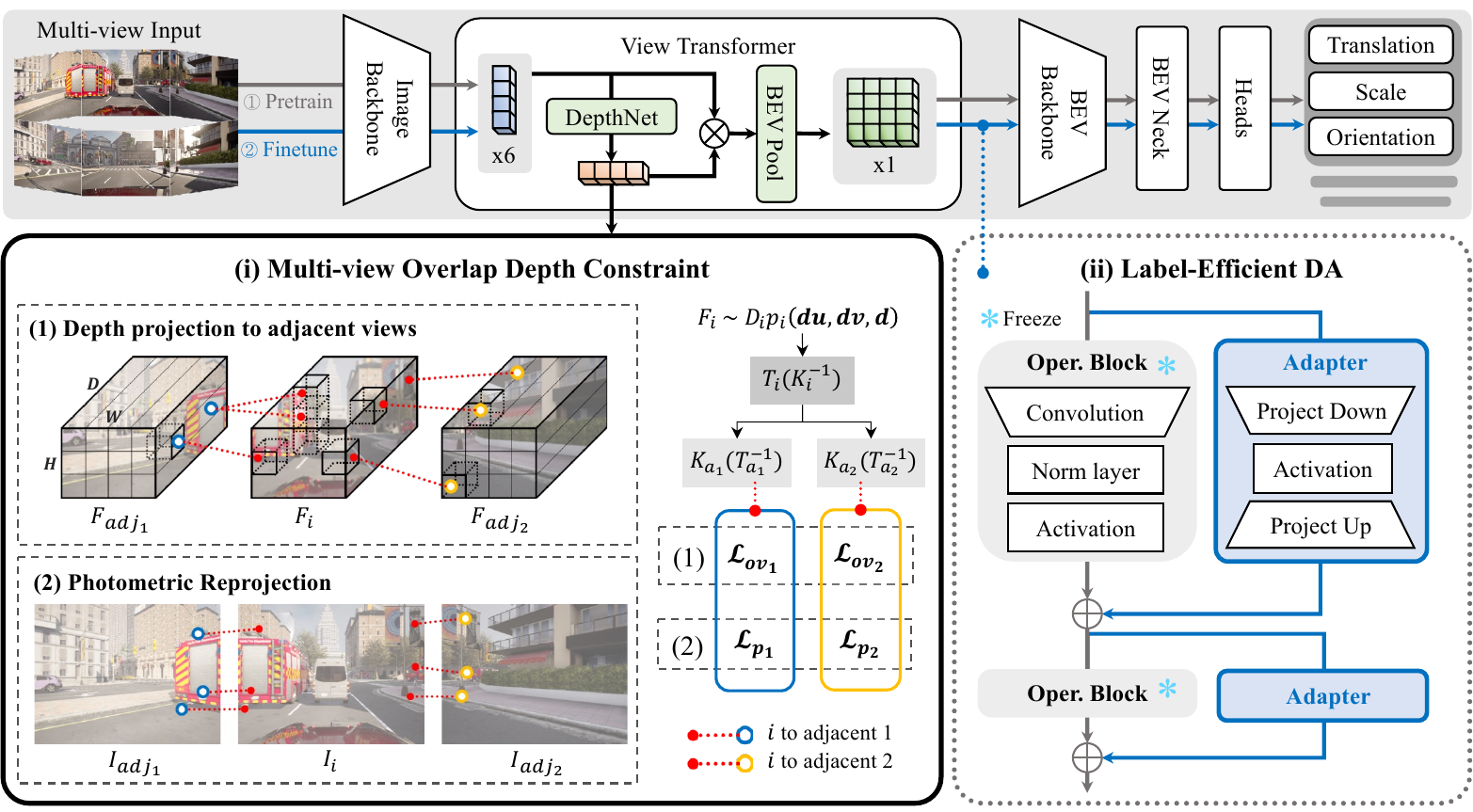}
    \caption{An overview of our proposed methodologies. Our proposed methods comprise two major parts: (i) Multi-view Overlap Depth Constraint and (ii) Label-Efficient Domain Adaptation (LEDA). In addition, our framework employs two phases (\ie, pre-training, and then fine-tuning). Note that we adopt our proposed depth constraint in both phases, and LEDA only in the fine-tuning phase.}
    \label{fig:main_architecture}
\end{figure}


\subsection{Domain Shifts in Multi-view 3D Object Detection}
\label{3.2}
In this section, we analyze and report \textit{de facto} domain shift problems arising in the Autonomous Driving system.
As shown in~\ref{3.1}, recent works adopt camera parameters $K$ and $T$ as extra inputs in addition to multi-view image $I$.
As reported by~\cite{wang2023towards}, assuming that the conditional distribution of outputs for given inputs, is the same across domains, it is explained that shifts in the domain distribution are caused by inconsistent marginal distributions of inputs.
To mitigate these issues, recent generalization approaches~\cite{wang2023towards, gu2021pit, li2022unsupervised, facil2019cam, klinghoffer2023towards} often focus on covariate shift in geometric feature representation mainly due to optical changes (\ie, Focal length, Field-of-View, and pixel size). 

This is the only part of a story.
We experience drastic performance drops (up to -54$\%$ / -67$\%$ performance drop in NDS and mAP, respectively, as shown in Fig~\ref{fig:domain_shift} (b)) from non-intrinsic factors (\ie, only extrinsic shifts).
Especially, we capture a phenomenon wherein the actual depth scale from an ego-vehicle's visual sensor to an object (Fig~\ref{fig:domain_shift} (a) red boxes) varies depending on the sensor's installation location.
Followed by Pythagorean theorem, as the height difference $\Delta h$ increases, the depth scale difference $\Delta d$ also increases accordingly.
Note that this is not limited to height solely; any shifts in deployment translation (\eg, along the x, y, or z axis) lead to changes in actual depth scale.
As a result, perspective view differences significantly hinder the model's three-dimensional geometric understanding by causing depth inconsistency.
To address above drawbacks, we introduce a novel penalising strategy that effectively boost depth consistency in various camera geometry shifts.


\subsection{Multi-view Overlap Depth Constraint}
\label{met:MODC}

\myparagraph{Motivation.}
Recently, previous efforts~\cite{zhao2021camera, wang2023towards, klinghoffer2023towards, wang2019perspective} augment multi-view images to generalize challenging perspective view gaps. However, these strategies suffer from poor generalizability in cross-domain scenarios, primarily due to the underestimated extent of view change between different sensor deployments as reported in section~\ref{3.2}.
To alleviate perspective gaps, we introduce Multi-view Overlap Depth Constraint, effectively encouraging perspective view-invariant learning.
Here, we start from three key assumptions:
First, perspective shifts between adjacent cameras in multi-view modalities are non-trivial and varied, closely akin to those observed in cross-domains (\eg, nuScenes $\rightarrow$ Lyft).
Second, visual odometry techniques such as Structure from Motion (SfM) and Simultaneous Localization and Mapping (SLAM) often benefit from improved depth consistency through relationships between adjacent views (\eg, relative pose estimation).
Third, in multi-view modalities, overlap regions serve as strong geometric triangular clues, seamlessly bridging between adjacent views.
However, under conditions where camera parameters are input, off-the-shelf pose estimation~\cite{godard2019digging, bian2023nope, lyu2021hr, zhou2021self, wei2022surround} leads to ambiguity in learning precise geometry.
To mitigate these issues, we introduce a novel depth constraint (Fig.~\ref{fig:main_architecture} (i)) with overlap regions between adjacent cameras.

\myparagraph{Approach.}
To achieve generalized BEV extraction, we directly constrain depth estimation network from adjacent overlap regions between multi-view cameras. 
Also, we advocate that multi-frame image inputs substantially complement geometric understanding in dynamic scenes with speedy translation and rotation shifts. 
To this end, we formulate corresponding depth $D^{\ast}$ leveraging spatial and temporal adjacent views.
First, we calculate overlap transformation matrices $T_{i \rightarrow j}$ from Eq.~\ref{eq:cor_depth}.
\begin{equation}
\label{eq:cor_depth}
    D_{i \rightarrow j}^{\ast}p_{i \rightarrow j}^{\ast} \sim K_j(T_j^{-1})T_i(K_i^{-1})D_ip_i,
\end{equation}
where $K$ and $T$ are the intrinsic and extrinsic camera parameters. $p_{i \rightarrow j}^{\ast}$ and $p_{i}$ denote corresponding pixels between adjacent views and $D$ represent depth prediction.
Then, we directly penalise unmatched corresponding depth $D^{\ast}$ to smoothly induce perspective-agnostic learning as follow Eq.~\ref{eq:loss_ov}
\begin{equation}
\label{eq:loss_ov}
    \mathcal{L}_{ov} = \sum_{(i,j)}{d(D_{j}, D_{i \rightarrow j}^{\ast})}, 
\end{equation}
where $d$ represents Euclidean Distance.
Also, we observe that the photometric reprojection error significantly alleviate relative geometric ambiguity. 
Especially, slow convergence may occur mainly due to incorrect relationships in small overlap region (about 30$\%$ of full resolution). 
To mitigates these concern, we effectively boost elaborate 2D matching, formulating $\mathcal{L}_p$ as follow Eq.~\ref{eq:loss_p}:
\begin{equation}
\label{eq:loss_p}
    \mathcal{L}_p = \sum_{(i,j)}pe(I_j \langle K_j,P_j \rangle, I_j \langle K_j, T_{i \rightarrow j}, P^*_{i \rightarrow j} \rangle),
\end{equation}
where $P$ represents point clouds generated by $D$, and $pe$ is photometric error by SSIM~\cite{wang2004image}. Also, $\langle \cdot \rangle$ denotes bilinear sampling on RGB images. Concretely, we take two advantages leveraging $\mathcal{L}_p$ in \textit{narrow occluded regions}; First, $\mathcal{L}_p$ effectively mitigates the triangular misalignment. Second, $\mathcal{L}_p$ potentially supports insufficiently scaled $\mathcal{L}_{ov}$. Ultimately, we alleviate perspective view gaps by directly constraining the corresponding depth and the photometric matching between adjacent views.


\subsection{Label-Efficient Domain Adaptation}
\label{met:UDGA}

\myparagraph{Motivation.}
There exist practical challenges in developing and deploying multi-view 3D object detectors for safety-critical self-driving vehicles. Each vehicle and each sensor requires its own model that can successfully operate in various conditions (\eg, dynamic weather, location, and time). Furthermore, while collecting large-scale labels in diverse environments is highly recommended, it is extremely expensive, inefficient and time-consuming. Among those, we are particularly motivated to tackle the following: (i) Stable performance, (ii) Efficiency of training, (iii) Preventing catastrophic forgetting, and (iv) Minimizing labeling cost. To satisfy these practical requirements, we carefully design an efficient and effective learning strategy, Label-Efficient Domain Adaptation (LEDA) that seamlessly transferring and preserving their own potentials leveraging a few annotated labels.

\myparagraph{Approach.}
In this paper, we propose Label-Efficient Domain Adaptation, a novel strategy to seamlessly bridge domain gaps leveraging a small amount of target data. 
To this end, we add extra parameters $\mathcal{A}$~\cite{houlsby2019parameter} consisting of bottleneck structures (\ie, projection down $\phi_{down}$ and up $\phi_{up}$ layers).
\begin{equation}
\label{eq:adapter}
    \mathcal{A}(x) = \phi_{up}(\sigma(\phi_{down}(BN(x)))),
\end{equation}
where $\sigma$ and $BN$ indicates activation function and batch normalization. 
We parallelly build $\mathcal{A}$ alongside pre-trained operation blocks $\mathcal{B}$ (\eg, convolution, and linear block) in Fig.~\ref{fig:main_architecture} (ii) and Eq.~\ref{eq:adaptation};
\begin{equation}
\label{eq:adaptation}
    y = \mathcal{B}(x) + \mathcal{A}(x),
\end{equation}
Firstly, we feed $x$ into $\phi_{down}$ to compress its shape to $[H/r, W/r]$, where $r$ is the rescale ratio, and then utilize $\phi_{up}$ to restore it to $[H, W]$. 
Secondly, we fuse each outputs from $\mathcal{B}$, and Adapter by exploiting skip-connections that directly link between the downsampling and upsampling paths.
By doing so, these extensible modules allow to capture high-resolution spatial details while reducing network and computational complexity.
Plus, it notes worthy that they are initialized by a near-identity function to preserve previously updated weights.
Finally, our frameworks lead to stable recognition in both source and target domains, incrementally adapting without forgetting pre-trained knowledge.


\subsection{Optimization Objective}
\label{met:loss_total}
In this section, we optimize our proposed framework UDGA using the total loss function $\mathcal{L}_{total}$ (as shown in Eq.~\ref{eq:total_loss}) during both phases (\ie, pre-train and fine-tune). $\mathcal{L}_{det}$ denotes the detection task loss.
\begin{equation}
\label{eq:total_loss}
    \mathcal{L}_{total} = \lambda_{det}\mathcal{L}_{det} + \lambda_{ov}\mathcal{L}_{ov} + \lambda_{p}\mathcal{L}_{p},
\end{equation}
where we grid-search $\lambda_{det}$, $\lambda_{ov}$ and $\lambda_{p}$ to harmonize $\mathcal{L}_{det}$, $\mathcal{L}_{ov}$ and $\mathcal{L}_{p}$. Specifically, $\mathcal{L}_{total}$ supervises $\mathcal{B}$ during generalization and $\mathcal{A}$ during adaptation, respectively. As a result, these strategies enable efficient learning of optimal representations in target domains while preserving pre-trained ones.

\section{Experimental Results}
\label{experiment}

\begin{table}[t!]
    \caption{Comparison of Domain Generalization performance with existing SOTA techniques. The \textbf{bold} values indicate the best performance. Note that all methods are evaluated on `car' category.}
    \label{tab:DG}
    \centering
    \resizebox{\linewidth}{!}{%
        \begin{tabular}{p{3.5cm}|p{3.5cm}|cccccc}
            \toprule
            Task & Method & ~NDS$^{\hat{*}}$$\uparrow$~ & ~mAP$\uparrow$~ & ~mATE$\downarrow$~ & ~mASE$\downarrow$~ & ~mAOE$\downarrow$~ & Closed Gap$\uparrow$\\
            \midrule
            \multirow{7}{*}{Lyft $\rightarrow$ nuScenes} & \textit{Oracle}  & 0.587 & 0.475 & 0.577 & 0.177 & 0.147 & \\\cmidrule{2-8}
            & \textit{Direct Transfer}  & 0.213 & 0.102 & 1.143 & 0.239 & 0.789 \\
            & CAM-Convs~\cite{facil2019cam}                         & 0.181 & 0.098 & 1.198 & 0.209 & 1.064 & -8.6$\%$\\
            & Single-DGOD~\cite{wu2022single}                       & 0.198 & 0.105 & 1.166 & 0.222 & 0.905 & -4.0$\%$\\
            & DG-BEV~\cite{wang2023towards}                            & 0.374 & 0.268 & 0.764 & 0.205 & 0.591 & +43.0$\%$\\
            & PD-BEV~\cite{lu2023towards}                            & 0.344 & 0.263 & \textbf{0.746} & 0.186 & 0.790 & +35.0$\%$\\
            \cmidrule{2-8}
            \coloredrowcell{DCDCDC}  & Ours & \textbf{0.421} & \textbf{0.281} & 0.759 & \textbf{0.183} & \textbf{0.377} & \textbf{+55.6}$\%$\\
            \midrule
            & \textit{Oracle} & 0.684 & 0.602 & 0.471 & 0.152 & 0.078 \\\cmidrule{2-8}
            \multirow{5}{*}{nuScenes $\rightarrow$ Lyft} & \textit{Direct Transfer} & 0.296 & 0.112 & 0.997 & 0.176 & 0.389 \\
            & CAM-Convs                            & 0.316 & 0.145 & 0.999 & 0.173 & 0.368 & +5.2$\%$\\
            & Single-DGOD                          & 0.332 & 0.159 & 0.949 & 0.174 & 0.358 & +9.3$\%$\\
            & DG-BEV                               & 0.437 & 0.287 & 0.771 & 0.170 & 0.302 & +36.3$\%$\\
            & PD-BEV                               & 0.458 & 0.304 & 0.709 & 0.169 & 0.289 & +41.8$\%$\\
            \cmidrule{2-8}
            \coloredrowcell{DCDCDC} & Ours & \textbf{0.487} & \textbf{0.324} & \textbf{0.709} & \textbf{0.162} & \textbf{0.180} & \textbf{+49.2}$\%$\\
            \midrule
            & \textit{Oracle} & 0.587 & 0.475 & 0.577 & 0.177 & 0.147 \\\cmidrule{2-8}
            \multirow{4}{*}{Waymo $\rightarrow$ nuScenes}  
            & \textit{Direct Transfer}          & 0.133 & 0.032 & 1.305 & 0.768 & 0.532 \\
            & CAM-Convs                         & 0.215 & 0.038 & 1.308 & 0.316 & 0.506 & +18.1$\%$\\
            & Single-DGOD                       & 0.007 & 0.014 & 1.000 & 1.000 & 1.000 & -27.8$\%$\\
            & DG-BEV                            & 0.472 & 0.303 & 0.689 & \textbf{0.218} & 0.171 & +74.7$\%$\\
            \cmidrule{2-8}
            \coloredrowcell{DCDCDC} & Ours & \textbf{0.477} & \textbf{0.326} & \textbf{0.684} & 0.263 & \textbf{0.168} & \textbf{+75.8}$\%$\\
            \midrule
            & \textit{Oracle}  & 0.649 & 0.552 & 0.528 & 0.148 & 0.085 \\\cmidrule{2-8}
            \multirow{4}{*}{nuScenes $\rightarrow$ Waymo} 
            & \textit{Direct Transfer}          & 0.178 & 0.040 & 1.303 & 0.265 & 0.790 \\
            & CAM-Convs                         & 0.185 & 0.045 & 1.301 & 0.253 & 0.773 & +1.5$\%$\\
            & Single-DGOD                       & 0.164 & 0.034 & 1.305 & 0.262 & 0.855 & -3.0$\%$\\
            & DG-BEV                            & 0.415 & 0.297 & 0.822 & \textbf{0.216} & 0.372 & +50.3$\%$\\
            \cmidrule{2-8}
            \coloredrowcell{DCDCDC} &  Ours & \textbf{0.459} & \textbf{0.349} & \textbf{0.754} & 0.289 & \textbf{0.250} & \textbf{+59.7}$\%$\\
            \bottomrule
        \end{tabular}}
\end{table}

In this section, we showcase the overall performance of our methodologies on landmark datasets for 3D Object Detection: Waymo~\cite{sun2020scalability}, Lyft~\cite{houston2021one}, and nuScenes~\cite{caesar2020nuscenes}. 
The three datasets have different specifications; thus, we convert them to a unified detection range and coordinates for accurate comparison. 
We also adopt only seven parameters to achieve consistent training results under the same conditions: the location of centerness $(x, y, z)$, the size of box $(l, w, h)$, and heading angle $\theta$. 
Additionally, we summarize 3D Object Detection datasets and implementation details in Appendix~\ref{apx:dataset}. 


\subsection{Evaluation Metric}
In this paper, following DG-BEV~\cite{wang2023towards} evaluation details, we adopt the alternative metric NDS$^{\hat{*}}$ (as shown in Eq.~\ref{eq:metric}) that aggregates mean Average Precision (mAP), mean Average Translation Error (mATE), mean Average Scale Error (mASE), and mean Average Orientation Error (mAOE).
\begin{equation}
\label{eq:metric}
\text{NDS}^{\hat{*}} = \frac{1}{6}[3\text{mAP} + \sum_{\text{mTP} \in \mathbb{TP}}(1-\text{min}(1, \text{mTP}))]
\end{equation}
We reconstruct the unified category for Unified Domain Generalization and Adaptation as follows:
the ‘car’ for nuScenes and Lyft, and the ‘vehicle’ for Waymo.
Furthermore, we only validate performance in the range of $x$, $y$ axis from -50m to 50m.
Note that we offer an empirical lower bound $\textbf{\textit{Direct Transfer}}$ (\ie, directly evaluating the model pre-trained in the source domain only), and an empirical upper bound $\textbf{\textit{Oracle}}$ (\ie, evaluating the model fully supervised in the target domain). 
We report $\textbf{Full F.T.}$ (\ie, fine-tuning all parameters from the pre-trained source model) and $\textbf{Adapter}$ (\ie, parameter efficient fine-tuning without our proposed depth constraint methods from the pre-trained source model)
Furthermore, we formulate \textbf{Closed Gap}-representing the hypothetical closed gap by
\begin{equation}
    \label{metric:closed_gap}
    \text{Closed Gap}=\frac{\text{NDS}_{\text{model}} - \text{NDS}_{\text{Direct Transfer}}}{\text{NDS}_{\text{Oracle}} - \text{NDS}_{\text{Direct Transfer}}} \times 100\%.
\end{equation}

\begin{table}[t]
    \caption{Comparison of UDGA performance on BEVDepth with various PEFT modules, SSF~\cite{liu2022few}, and Adapter~\cite{houlsby2019parameter}. We construct six different target data splits from 1$\%$ to 100$\%$. 
    Additionally, $\#$ Params denote the number of parameters for training. Note that --- represents \textit{`Do not support'}.}
    \label{tab:UDGA}
    \centering
    \resizebox{\linewidth}{!}{
    \begin{tabular}{p{2.5cm}p{1.6cm}p{1.4cm}cccccc}
        \toprule 
        \multirow{2.3}{*}{Task} & \multirow{2.3}{*}{Method} & \multirow{2.3}{*}{$\#$ Params} & \multicolumn{6}{c}{NDS$^{\hat{*}}$$\uparrow$ / mAP$\uparrow$} \\
        \cmidrule{4-9}
        \multirow{10}{*}{Lyft $\rightarrow$ nuScenes} & & & 1$\%$ & 5$\%$
        & 10$\%$ & 25$\%$ & 50$\%$ & 100$\%$  \\
        \midrule
        & \textit{Oracle} & 51.7M & --- & --- & --- & --- & --- & 0.587 / 0.475 \\\cmidrule{2-9}
        & Full FT & 51.7M & 0.476 / 0.369 & 0.515 / 0.434 & 0.547 / 0.434 & 0.577 / 0.464 & 0.590 / 0.483 & 0.610 / 0.506 \\
        & SSF~\cite{liu2022few} & 1M & 0.245 / 0.079 & 0.294 / 0.112 & 0.360 / 0.256 & 0.374 / 0.266  & 0.421 / 0.327 & 0.439 / 0.275 \\
        & Adapter-B & 21.3M & 0.465 / 0.283 & 0.481 / 0.365 & 0.511 / 0.384 & 0.558 / 0.444 & 0.569 / 0.460 & 0.581 / 0.473 \\
        & Adapter-S & 8.8M & 0.326 / 0.134 & 0.372 / 0.161 & 0.444 / 0.255 & 0.465 / 0.283 & 0.509 / 0.390 & 0.538 / 0.443 \\
        \cmidrule{2-9}
        \coloredrowcell{DCDCDC} & Ours & 8.8M & \textbf{0.526} / \textbf{0.404} & \textbf{0.563} / \textbf{0.444} & \textbf{0.573} / \textbf{0.457} & \textbf{0.592} / \textbf{0.481} & \textbf{0.609} / \textbf{0.510} & \textbf{0.614} / \textbf{0.507} \\
        
        \midrule
        \multirow{7}{*}{nuScenes $\rightarrow$ Lyft} 
        & \textit{Oracle} & 51.7M & --- & --- & --- & --- & --- & 0.684 / 0.602 \\\cmidrule{2-9}
        & Full FT & 51.7M & 0.531 / 0.390 & 0.594 / 0.473 & 0.623 / 0.513 & 0.650 / 0.549 & 0.678 / 0.587 & 0.700 / 0.615 \\
        & SSF & 1M & 0.316 / 0.115 & 0.355 / 0.145 & 0.386 / 0.185 & 0.420 / 0.230 & 0.447 / 0.269 & 0.470 / 0.300 \\
        & Adapter-B & 21.3M & 0.499 / 0.328 & 0.556 / 0.465 & 0.584 / 0.475 & 0.633 / 0.532 & 0.670 / 0.564 & 0.684 / 0.596 \\
        & Adapter-S & 8.8M & 0.420 / 0.230 & 0.463 / 0.325 & 0.500 / 0.356 & 0.537 / 0.400 & 0.561 / 0.426 & 0.573 / 0.442 \\\cmidrule{2-9}
        \coloredrowcell{DCDCDC} & Ours & 8.8M & \textbf{0.578} / \textbf{0.462} & \textbf{0.613} / \textbf{0.506} & \textbf{0.638} / \textbf{0.537} & \textbf{0.665} / \textbf{0.572} & \textbf{0.675} / \textbf{0.586} & \textbf{0.706} / \textbf{0.626} \\
        \bottomrule
    \end{tabular}}
\end{table}

\begin{figure}[t!]
    \centering
    \includegraphics[width=1\linewidth]{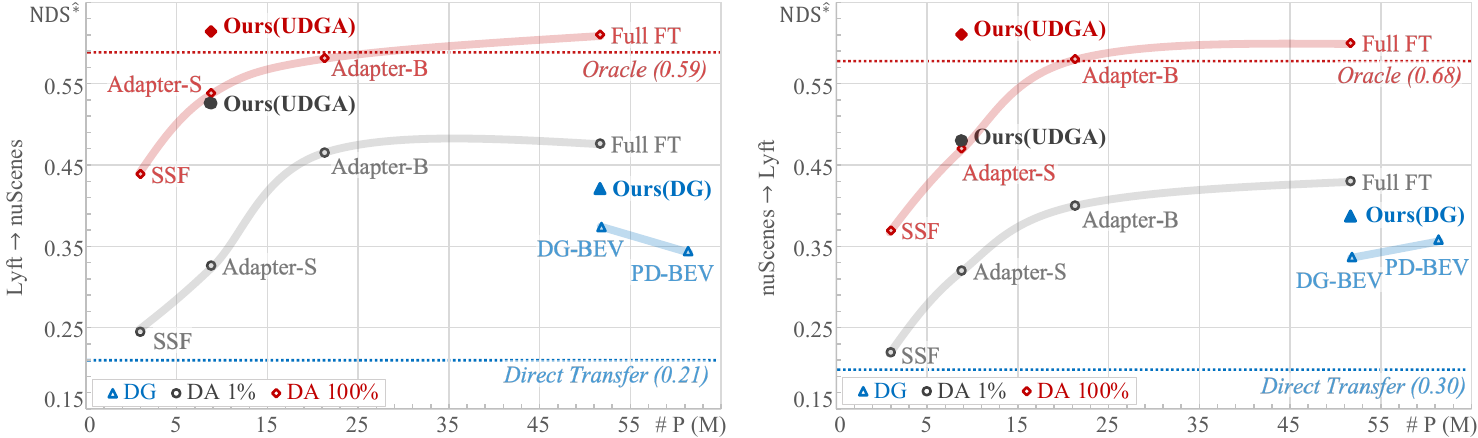}
    \caption{
    Performance relative to training parameters. The Domain Generalization task is represented in blue, while the Domain Adaptation task is divided into two stages: 1$\%$ in gray and 100$\%$ in red.}
   
    \label{fig:reb_LEDA}
\end{figure}


\subsection{Experiment Results}
\myparagraph{Performance Comparison in Domain Generalization.}
As shown in Tab.~\ref{tab:DG}, we showcase four challenging generalization scenarios, and quantitatively compare our proposed methodology with existing state-of-the-art methods, which include CAM-Conv~\cite{facil2019cam}, Single-DGOD~\cite{wu2022single}, DG-BEV~\cite{wang2023towards}, and PD-BEV~\cite{lu2023towards}.
Here, we observe that these methods still struggle to fully pilot geometric shifts from perspective changes in cross-domain scenarios.
Importantly, in Lyft $\rightarrow$ nuScenes, existing methods suffer from the orientation error mainly due to significantly different ground truth directions (\ie, only recovering 0.198 mAOE).
In nuScenes $\rightarrow$ Waymo (\ie, one of the most challenging scenarios due to the rear camera drop), previous approaches still show a significant gap compared to \textit{Oracle} (\ie, -49.7$\%$ Closed Gap).
In this paper, our novel depth constraint notably addresses these issues, outperforming existing SOTAs (especially, up to +4.7$\%$ NDS and +12.6$\%$ Closed Gap better than DG-BEV in Lyft $\rightarrow$ nuScenes).
Especially, leveraging triangular clues to explicitly supervise occluded depth contributes significantly to improving geometric consistency compared to prior approaches~\cite{wang2023towards, lu2023towards, wu2022single, facil2019cam}.
Overall, we demonstrate that our novel approaches significantly enhance perspective-invariance, featuring strong association in occluded regions between multi-views.

\myparagraph{Performance Comparison in UDGA.}
In Tab.~\ref{tab:UDGA}, we show that our proposed Unified Domain Generalization and Adaptation performance compared with various PEFT approaches (\ie, SSF~\cite{liu2022few}, and Adapter~\cite{houlsby2019parameter}).
SSF directly scale and shift the deep features extracted by pre-trained operation blocks, leveraging additional normalization parameters.
Adapter represents sole module performance without our proposed constraint; Adapter-B, and Adapter-S denotes base, and small version, respectively.

Existing PEFT paradigms benefit from fine-tuning only extra parameters, retaining previously updated weights.
However, we observe that these paradigms do not successfully adapt to the covariate shifts originated by challenging geometric differences as reported in section~\ref{3.2}.
More specifically, SSF and Adapter-S, which exploit a small number of parameters, begin to capture transferable representations and then marginally adapt at the 10$\%$ data split.
Also, Adapter-B leveraging 21.3M parameters provide poor adaptation capability (\ie, inferior to Scratch and Full FT in Lyft $\rightarrow$ nuScenes 100$\%$).

However, our proposed strategy seamlessly adapt to target domains in 1$\%$, and 5$\%$, effectively bridge perspective gaps.
Furthermore, our proposed strategy show superior performance gain (outperforming Scratch in Lyft $\rightarrow$ nuScenes 50$\%$, and Full FT in both Lyft $\rightarrow$ nuScenes, and nuScenes $\rightarrow$ Lyft 100$\%$), effectively adapting to novel targets.
It is noteworthy that the most effective adaptation is achieved by updating extra parameters (less than 20$\%$ of the total), which demonstrates the practicality and efficiency of our novel UDGA strategy as shown in Fig.~\ref{fig:reb_LEDA}.
In addition, unlike Full FT, it proves that our UDGA framework stably adapts to the target without forgetting previously learned knowledge as shown in Fig.~\ref{fig:gda_graph} and Tab.~\ref{tab:UDGA2}.
Overall, our proposed method demonstrate the effectiveness of training strategy in various experimental setups, efficiently expanding to targets with about 20$\%$ of overall parameters.
Note that we report additional experiments and details of UDGA in Appendix~\ref{apx:add_exp}.

\begin{table}[t!]
    \caption{Ablation studies on UDGA (10$\%$ Adaptation). $\mathcal{B}$ and $\mathcal{A}$ represents pre-trained blocks and LEDA blocks, respectively. Note that we train $\mathcal{B}$ and $\mathcal{A}$, alternatively (\ie, pre-train and fine-tune).}
    \label{exp:ablation}
    \centering
    \resizebox{\linewidth}{!}{%
    \begin{tabular}{>{\centering\arraybackslash}p{1.2cm} >{\centering\arraybackslash}p{1.2cm} >{\centering\arraybackslash}p{1.2cm}|>{\centering\arraybackslash}p{1.2cm} >{\centering\arraybackslash}p{1.2cm} >{\centering\arraybackslash}p{1.2cm}|>{\centering\arraybackslash}p{1.5cm} >{\centering\arraybackslash}p{1.5cm}|>{\centering\arraybackslash}p{1.5cm} >{\centering\arraybackslash}p{1.5cm}}
        \toprule 
        \multicolumn{3}{c}{Pre-train $\mathcal{B}$ (100$\%$ source)} & \multicolumn{3}{|c|}{Fine-tune $\mathcal{A}$ (10$\%$ target)} & \multicolumn{2}{c|}{Lyft $\rightarrow$ nuScenes} & \multicolumn{2}{c}{nuScenes $\rightarrow$ Lyft}  \\
        \cmidrule{1-10}
        $\mathcal{L}_{det}$ & $\mathcal{L}_{ov}$ & $\mathcal{L}_{p}$ & 
        $\mathcal{L}_{det}$ & $\mathcal{L}_{ov}$ & $\mathcal{L}_{p}$ & ~NDS$^{\hat{*}}$$\uparrow$~ & ~mAP$\uparrow$~ & ~NDS$^{\hat{*}}$$\uparrow$~ & ~mAP$\uparrow$~\\
        \midrule                                                                      
        \checkmark &&&& &                                                           & 0.213 & 0.102 & 0.296 & 0.112  \\ 
        \checkmark & \checkmark &&& &                                               & 0.403 & 0.262 & 0.485 & 0.323  \\ 
        \rowcolor[HTML]{DCDCDC}\checkmark & \checkmark & \checkmark && &            & 0.421 & 0.281 & 0.488 & 0.309  \\ 
        \cmidrule{1-10}
        \checkmark &&& \checkmark & &                                               & 0.444 & 0.255 & 0.500 & 0.356  \\
        \checkmark & \checkmark & \checkmark & \checkmark & &                       & 0.516 & 0.407 & 0.590 & 0.482  \\
        \checkmark & \checkmark & \checkmark & \checkmark & \checkmark &            & 0.552 & 0.441 & 0.632 & 0.530  \\
        \rowcolor[HTML]{DCDCDC}\checkmark & \checkmark & \checkmark & \checkmark & \checkmark & \checkmark & \textbf{0.638} & \textbf{0.537} & \textbf{0.573} & \textbf{0.457}  \\        
        \bottomrule
    \end{tabular}}
\end{table}

\begin{table}[t!]
    \caption{Ablation studies on Domain Generalization with our novel depth constraint modules, $\mathcal{L}_{ov}$ and $\mathcal{L}_{p}$. Lidar and SS each represents LiDAR depth supervision and Self-Supervised overlap depth. 
    }
    \label{tab:ab_depth}
    \centering
    \resizebox{\linewidth}{!}{
    \begin{tabular}{p{3.2cm}|>{\centering\arraybackslash}p{1cm} >{\centering\arraybackslash}p{1cm}|p{3.3cm}|ccccc}
        \toprule 
        Task & Lidar & SS & \multicolumn{1}{c}{Method} & ~NDS$^{\hat{*}}$$\uparrow$~ & ~mAP$\uparrow$~ & ~mATE$\downarrow$~ & ~mASE$\downarrow$~ & ~mAOE$\downarrow$~ \\
        \midrule
        \multirow{5.5}{*}{Lyft $\rightarrow$ nuScenes} 
        & $\checkmark$ & & $\mathcal{L}_{d}$ & 0.213 & 0.102 & 1.143 & 0.239 & 0.789 \\ 
        & $\checkmark$ & $\checkmark$ & $\mathcal{L}_{d}$ + $\mathcal{L}_{ov}$ + $\mathcal{L}_{p}$ & 0.396 & 0.266 & 0.758 & 0.172 & 0.495 \\
        \cmidrule{2-9}
        & & $\checkmark$ & $\mathcal{L}_{ov}$ & 0.403 & 0.262 & 0.757 & 0.183 & 0.426 \\
        & & $\checkmark$ & $\mathcal{L}_{ov}$ + $\mathcal{L}_{p}$ & 0.407 & 0.265 & \textbf{0.747} & \textbf{0.179} & 0.428 \\        
        \coloredrowcell{DCDCDC} & & $\checkmark$ & $\mathcal{L}_{ov}$ + $\mathcal{L}_{p}$ + \textit{ext~aug} & \textbf{0.421} & \textbf{0.281} & 0.759 & 0.183 & \textbf{0.377} \\
        \midrule
        \multirow{5.5}{*}{nuScenes $\rightarrow$ Lyft} 
        & $\checkmark$ & & $\mathcal{L}_{d}$ & 0.296 & 0.112 & 0.997 & 0.176 & 0.389 \\
        & $\checkmark$ & $\checkmark$ & $\mathcal{L}_{d}$ + $\mathcal{L}_{ov}$ + $\mathcal{L}_{p}$ & 0.483 & 0.327 & 0.718 & 0.163 & 0.204 \\
        \cmidrule{2-9}        
        & & $\checkmark$ & $\mathcal{L}_{ov}$                     & 0.485 & 0.323 & 0.731 & \textbf{0.161} & 0.171 \\
        & & $\checkmark$ & $\mathcal{L}_{ov}$ + $\mathcal{L}_{p}$ & 0.487 & \textbf{0.324} & 0.709 & 0.162 & 0.180 \\
        \coloredrowcell{DCDCDC} & & $\checkmark$ & $\mathcal{L}_{ov}$ + $\mathcal{L}_{p} + \textit{ext~aug}$ 
                                                   & \textbf{0.488} & 0.309 & \textbf{0.705} & 0.169 & \textbf{0.123} \\
        \bottomrule
    \end{tabular}}
\end{table}


\subsection{Ablation studies}
\myparagraph{Exploring the Synergy Between Modules.}
To better understand the role of each module, we present ablation studies of UDGA in this experiment (Tab.~\ref{exp:ablation}). Precisely, we aim to analyze the pros and cons in both training steps (\ie, pre-train $\mathcal{B}$ and fine-tune $\mathcal{A}$), with the objective of effectively elucidating the plausibility of UDGA. First, the strategy trained from scratch leveraging our depth constraint significantly recovers performance drop from the sensor deployment shift (up to +20.8$\%$ NDS). However, this strategy finds it difficult to provide a practical solution for Multi-view 3DOD, mainly due to unsatisfying generalizability. Additionally, although LEDA without $\mathcal{L}_{ov}$ and $\mathcal{L}_{p}$ yields improved performance, it fails to transfer its previously learned potential, resulting in only +2.3$\%$ NDS compared to our individual depth constraint. To tackle these issues, we concentrate on bridging two distinct domains by capturing generalized perspective features. Especially, our depth constraint (only trained during pre-training $\mathcal{B}$) significantly encourages understanding of the target in LEDA during fine-tuning $\mathcal{A}$ with a 10$\%$ split, addressing the geometric covariate shift (+30.3$\%$ NDS). Furthermore, UDGA strategy using $\mathcal{L}_{ov}$ and $\mathcal{L}_{p}$ in both phases learns the transferable knowledge and shows impressive improvement (+42.5$\%$ NDS). Finally, UDGA successfully presents an effective and efficient paradigm for Multi-view 3DOD, highlighting notable recovery in novel target scenarios.

\myparagraph{Effect of Overlap Depth Constraint.}
In Tab.~\ref{tab:ab_depth}, we carefully evaluate our depth constraint components in various cross-domain environments.
Here, $\mathcal{L}_d$ denotes depth supervision by LiDAR.
Also, we design \textit{ext aug} that globally rotate ground truths with randomly initialized angle $\alpha$ to release the direction shift.
More importantly, we observe that perspective view shifts from different sensor deployments lead to severe translation and orientation errors.
To tackle these issues, we advocate that $\mathcal{L}_{ov}$, which leverages strong relationships between adjacent views, effectively alleviating perspective gaps compared to $\mathcal{L}_d$ (recovering up to +19$\%$ NDS in Lyft $\rightarrow$ nuScenes).
$\mathcal{L}_p$ relieves slight misalignment, encouraging depth-scale consistency.
Additionally, our \textit{ext aug} substantially boost stable generalization, suppressing orientation errors (up to +1.4$\%$ additional NDS gain).
Consequently, our novel objectives ($\mathcal{L}_{ov}$ and $\mathcal{L}_{p}$) demonstrate their effectiveness, significantly tackling geometric errors.


\subsection{Qualitative Analysis}
\begin{figure}
    \centering
    \includegraphics[width=1\linewidth]{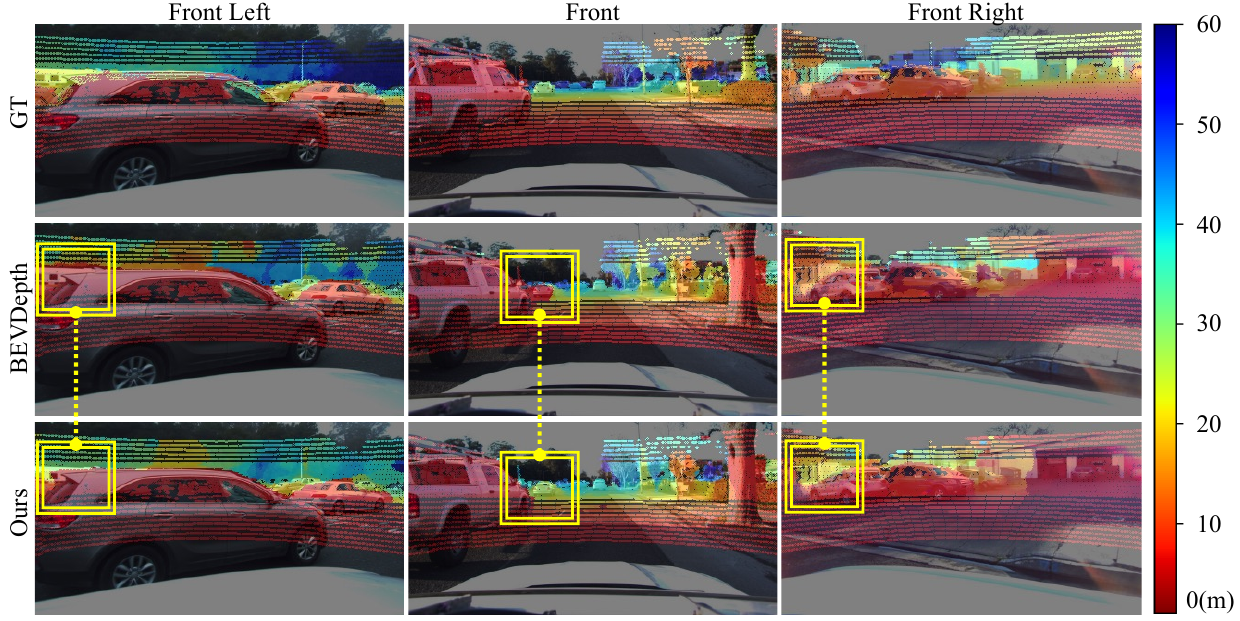}
    \caption{
        Qualitative depth visualizations of front view lineups in Lyft. The top row illustrates sparse depth ground truths projected from LiDAR point clouds. 
        The middle and bottom rows are the qualitative results of BEVDepth and Ours, respectively. Yellow boxes highlight the improved depth.
    }
    \label{fig:depth_qual}
\end{figure}

To qualitatively analyze the effectiveness of Multi-view Overlap Depth Constraint, we present additional visualized results in Fig.~\ref{fig:depth_qual}.
For accurate comparison, we conduct binary masking leveraging given sparse depth ground truths. 
In middle row, BEVDepth fail to perceive hard samples (\eg, far distant and occluded objects) in yellow boxes, mainly due to different extent of deformation relative to perspective as reported in section~\ref{3.2}.
We aim to tackle this problem, explicitly bridging adjacent views in various dynamic scenes.
Precisely, in bottom row, we showcase distinguishable results in yellow boxes, capturing semantic details from various view deformation. 
As as results, we qualitatively demonstrate that our proposed method effectively encourage depth consistency and detection robustness, significantly improving geometric understanding in cross-domain scenarios.

\section{Conclusion}
\label{conclusion}
\myparagraph{Limitations.}
While our work significantly improves the adaptability of 3D object detection, it cannot guarantee seamless adaptation due to several limitations, including:
(1) The performance does not match that of 3D object detection models using LiDAR point clouds.
(2) Our Multi-view Overlap Depth Constraint relies on the presence of overlapping regions between images.
(3) Achieving fully domain-agnostic approaches without any target labels remains challenging. As a
result, it is essential to incorporate a fallback plan when deploying the framework in safety-critical
real-world scenarios.

\myparagraph{Summary.}
Multi-View 3DOD models often face challenges in expanding appropriately to unfamiliar datasets due to inevitable domain shifts (\ie, changes in the distribution of data between the training and testing phases).
Especially, the limited resource (\eg, excessive computational overhead and taxing expensive and taxing data cost) leads to hinder the successful deployment of Multi-View 3DOD.
To mitigate above drawbacks, we carefully design Unified Domain Generalization and Adaptation (UDGA), a practical solution for developing Multi-View 3DOD.
We first introduce Multi-view Overlap Depth Constraint that advocates strong triangular clues between adjacent views, significantly bridging perspective gaps.
Additionally, we present a Label-Efficient Domain Adaptation approach that enables practical adaptation to novel targets with largely limited labels (\ie, 1$\%$ and 5$\%)$ without forgetting well-aligned source potential.
Our UDGA paradigm efficiently fine-tune additional parameters leveraging significantly fewer annotations by effectively transferring from the source to target domain.
In summary, our extensive experiments in various landmark datasets(\eg, nuScenes, Lyft and Waymo) show that our novel paradigm, UDGA, provide a practical solution, outperforming current state-of-the-art models on Multi-view 3D object detection.

\begin{ack}
This work was primarily supported by Samsung Advanced Institute of Technology (SAIT) (85\%), partially supported by Institute of Information \& communications Technology Planning \& Evaluation (IITP) grant funded by the Korea government (MSIT) (No. 2019-0-00079, Artificial Intelligence Graduate School Program (Korea University), 5\%), and Culture, Sports and Tourism R\&D Program through the Korea Creative Content Agency grant funded by the Ministry of Culture, Sports and Tourism in 2024(International Collaborative Research and Global Talent Development for the Development of Copyright Management and Protection Technologies for Generative AI, RS-2024-00345025, 5\%;
Development of technology for dataset copyright of multimodal generative AI model, RS-2024-00333068, 5\%).
\end{ack}

\bibliography{main}

\begin{thebibliography}{64}
\providecommand{\natexlab}[1]{#1}
\providecommand{\url}[1]{\texttt{#1}}
\expandafter\ifx\csname urlstyle\endcsname\relax
  \providecommand{\doi}[1]{doi: #1}\else
  \providecommand{\doi}{doi: \begingroup \urlstyle{rm}\Url}\fi

\bibitem[Bai et~al.(2022)Bai, Hu, Zhu, Huang, Chen, Fu, and Tai]{bai2022transfusion}
Xuyang Bai, Zeyu Hu, Xinge Zhu, Qingqiu Huang, Yilun Chen, Hongbo Fu, and Chiew-Lan Tai.
\newblock Transfusion: Robust lidar-camera fusion for 3d object detection with transformers.
\newblock In \emph{Proceedings of the IEEE/CVF conference on computer vision and pattern recognition}, pages 1090--1099, 2022.

\bibitem[Chang et~al.(2024)Chang, Roh, Jang, Lee, Ji, Oh, Park, Kim, and Kim]{chang2024cmda}
Gyusam Chang, Wonseok Roh, Sujin Jang, Dongwook Lee, Daehyun Ji, Gyeongrok Oh, Jinsun Park, Jinkyu Kim, and Sangpil Kim.
\newblock Cmda: Cross-modal and domain adversarial adaptation for lidar-based 3d object detection.
\newblock In \emph{Proceedings of the AAAI Conference on Artificial Intelligence}, volume~38, pages 972--980, 2024.

\bibitem[Liang et~al.(2022)Liang, Xie, Yu, Xia, Lin, Wang, Tang, Wang, and Tang]{liang2022bevfusion}
Tingting Liang, Hongwei Xie, Kaicheng Yu, Zhongyu Xia, Zhiwei Lin, Yongtao Wang, Tao Tang, Bing Wang, and Zhi Tang.
\newblock Bevfusion: A simple and robust lidar-camera fusion framework.
\newblock \emph{Advances in Neural Information Processing Systems}, 35:\penalty0 10421--10434, 2022.

\bibitem[Liu et~al.(2023)Liu, Tang, Amini, Yang, Mao, Rus, and Han]{liu2023bevfusion}
Zhijian Liu, Haotian Tang, Alexander Amini, Xinyu Yang, Huizi Mao, Daniela~L Rus, and Song Han.
\newblock Bevfusion: Multi-task multi-sensor fusion with unified bird's-eye view representation.
\newblock In \emph{2023 IEEE international conference on robotics and automation (ICRA)}, pages 2774--2781. IEEE, 2023.

\bibitem[Sun et~al.(2020)Sun, Kretzschmar, Dotiwalla, Chouard, Patnaik, Tsui, Guo, Zhou, Chai, Caine, et~al.]{sun2020scalability}
Pei Sun, Henrik Kretzschmar, Xerxes Dotiwalla, Aurelien Chouard, Vijaysai Patnaik, Paul Tsui, James Guo, Yin Zhou, Yuning Chai, Benjamin Caine, et~al.
\newblock Scalability in perception for autonomous driving: Waymo open dataset.
\newblock In \emph{Proceedings of the IEEE/CVF conference on computer vision and pattern recognition}, pages 2446--2454, 2020.

\bibitem[Caesar et~al.(2020)Caesar, Bankiti, Lang, Vora, Liong, Xu, Krishnan, Pan, Baldan, and Beijbom]{caesar2020nuscenes}
Holger Caesar, Varun Bankiti, Alex~H Lang, Sourabh Vora, Venice~Erin Liong, Qiang Xu, Anush Krishnan, Yu~Pan, Giancarlo Baldan, and Oscar Beijbom.
\newblock nuscenes: A multimodal dataset for autonomous driving.
\newblock In \emph{Proceedings of the IEEE/CVF conference on computer vision and pattern recognition}, pages 11621--11631, 2020.

\bibitem[Houston et~al.(2021)Houston, Zuidhof, Bergamini, Ye, Chen, Jain, Omari, Iglovikov, and Ondruska]{houston2021one}
John Houston, Guido Zuidhof, Luca Bergamini, Yawei Ye, Long Chen, Ashesh Jain, Sammy Omari, Vladimir Iglovikov, and Peter Ondruska.
\newblock One thousand and one hours: Self-driving motion prediction dataset.
\newblock In \emph{Conference on Robot Learning}, pages 409--418. PMLR, 2021.

\bibitem[Wang et~al.(2022)Wang, Guizilini, Zhang, Wang, Zhao, and Solomon]{wang2022detr3d}
Yue Wang, Vitor~Campagnolo Guizilini, Tianyuan Zhang, Yilun Wang, Hang Zhao, and Justin Solomon.
\newblock Detr3d: 3d object detection from multi-view images via 3d-to-2d queries.
\newblock In \emph{Conference on Robot Learning}, pages 180--191. PMLR, 2022.

\bibitem[Li et~al.(2022{\natexlab{a}})Li, Wang, Li, Xie, Sima, Lu, Qiao, and Dai]{li2022bevformer}
Zhiqi Li, Wenhai Wang, Hongyang Li, Enze Xie, Chonghao Sima, Tong Lu, Yu~Qiao, and Jifeng Dai.
\newblock Bevformer: Learning bird’s-eye-view representation from multi-camera images via spatiotemporal transformers.
\newblock In \emph{European conference on computer vision}, pages 1--18. Springer, 2022{\natexlab{a}}.

\bibitem[Roh et~al.(2022)Roh, Chang, Moon, Nam, Kim, Kim, Kim, and Kim]{roh2022ora3d}
Wonseok Roh, Gyusam Chang, Seokha Moon, Giljoo Nam, Chanyoung Kim, Younghyun Kim, Jinkyu Kim, and Sangpil Kim.
\newblock Ora3d: Overlap region aware multi-view 3d object detection.
\newblock \emph{arXiv preprint arXiv:2207.00865}, 2022.

\bibitem[Huang et~al.(2021)Huang, Huang, Zhu, Ye, and Du]{huang2021bevdet}
Junjie Huang, Guan Huang, Zheng Zhu, Yun Ye, and Dalong Du.
\newblock Bevdet: High-performance multi-camera 3d object detection in bird-eye-view.
\newblock \emph{arXiv preprint arXiv:2112.11790}, 2021.

\bibitem[Li et~al.(2023)Li, Ge, Yu, Yang, Wang, Shi, Sun, and Li]{li2023bevdepth}
Yinhao Li, Zheng Ge, Guanyi Yu, Jinrong Yang, Zengran Wang, Yukang Shi, Jianjian Sun, and Zeming Li.
\newblock Bevdepth: Acquisition of reliable depth for multi-view 3d object detection.
\newblock In \emph{Proceedings of the AAAI Conference on Artificial Intelligence}, volume~37, pages 1477--1485, 2023.

\bibitem[Facil et~al.(2019)Facil, Ummenhofer, Zhou, Montesano, Brox, and Civera]{facil2019cam}
Jose~M Facil, Benjamin Ummenhofer, Huizhong Zhou, Luis Montesano, Thomas Brox, and Javier Civera.
\newblock Cam-convs: Camera-aware multi-scale convolutions for single-view depth.
\newblock In \emph{Proceedings of the IEEE/CVF Conference on Computer Vision and Pattern Recognition}, pages 11826--11835, 2019.

\bibitem[Wang et~al.(2023)Wang, Zhao, Xu, Chen, Yu, Chang, Yang, and Zhao]{wang2023towards}
Shuo Wang, Xinhai Zhao, Hai-Ming Xu, Zehui Chen, Dameng Yu, Jiahao Chang, Zhen Yang, and Feng Zhao.
\newblock Towards domain generalization for multi-view 3d object detection in bird-eye-view.
\newblock In \emph{Proceedings of the IEEE/CVF Conference on Computer Vision and Pattern Recognition}, pages 13333--13342, 2023.

\bibitem[Lu et~al.(2023)Lu, Zhang, Lian, Du, and Chen]{lu2023towards}
Hao Lu, Yunpeng Zhang, Qing Lian, Dalong Du, and Yingcong Chen.
\newblock Towards generalizable multi-camera 3d object detection via perspective debiasing.
\newblock \emph{arXiv preprint arXiv:2310.11346}, 2023.

\bibitem[Hu et~al.(2021)Hu, Shen, Wallis, Allen-Zhu, Li, Wang, Wang, and Chen]{hu2021lora}
Edward~J Hu, Yelong Shen, Phillip Wallis, Zeyuan Allen-Zhu, Yuanzhi Li, Shean Wang, Lu~Wang, and Weizhu Chen.
\newblock Lora: Low-rank adaptation of large language models.
\newblock \emph{arXiv preprint arXiv:2106.09685}, 2021.

\bibitem[Lester et~al.(2021)Lester, Al-Rfou, and Constant]{lester2021power}
Brian Lester, Rami Al-Rfou, and Noah Constant.
\newblock The power of scale for parameter-efficient prompt tuning.
\newblock \emph{arXiv preprint arXiv:2104.08691}, 2021.

\bibitem[Liu et~al.(2021)Liu, Ji, Fu, Tam, Du, Yang, and Tang]{liu2021p}
Xiao Liu, Kaixuan Ji, Yicheng Fu, Weng~Lam Tam, Zhengxiao Du, Zhilin Yang, and Jie Tang.
\newblock P-tuning v2: Prompt tuning can be comparable to fine-tuning universally across scales and tasks.
\newblock \emph{arXiv preprint arXiv:2110.07602}, 2021.

\bibitem[Yan et~al.(2018)Yan, Mao, and Li]{yan2018second}
Yan Yan, Yuxing Mao, and Bo~Li.
\newblock Second: Sparsely embedded convolutional detection.
\newblock \emph{Sensors}, 18\penalty0 (10):\penalty0 3337, 2018.

\bibitem[Park et~al.(2021)Park, Ambrus, Guizilini, Li, and Gaidon]{park2021dd3d}
Dennis Park, Rares Ambrus, Vitor Guizilini, Jie Li, and Adrien Gaidon.
\newblock Is pseudo-lidar needed for monocular 3d object detection?
\newblock In \emph{IEEE/CVF International Conference on Computer Vision (ICCV)}, 2021.

\bibitem[Wang et~al.(2021{\natexlab{a}})Wang, Zhu, Pang, and Lin]{wang2021pgd}
Tai Wang, Xinge Zhu, Jiangmiao Pang, and Dahua Lin.
\newblock {Probabilistic and Geometric Depth: Detecting} objects in perspective.
\newblock In \emph{Conference on Robot Learning (CoRL) 2021}, 2021{\natexlab{a}}.

\bibitem[Wang et~al.(2021{\natexlab{b}})Wang, Zhu, Pang, and Lin]{wang2021fcos3d}
Tai Wang, Xinge Zhu, Jiangmiao Pang, and Dahua Lin.
\newblock Fcos3d: Fully convolutional one-stage monocular 3d object detection.
\newblock In \emph{Proceedings of the IEEE/CVF International Conference on Computer Vision}, pages 913--922, 2021{\natexlab{b}}.

\bibitem[Brazil and Liu(2019)]{brazil2019m3d}
Garrick Brazil and Xiaoming Liu.
\newblock M3d-rpn: Monocular 3d region proposal network for object detection.
\newblock In \emph{Proceedings of the IEEE/CVF International Conference on Computer Vision}, pages 9287--9296, 2019.

\bibitem[Liu et~al.(2020)Liu, Wu, and T{\'o}th]{liu2020smoke}
Zechen Liu, Zizhang Wu, and Roland T{\'o}th.
\newblock Smoke: Single-stage monocular 3d object detection via keypoint estimation.
\newblock In \emph{Proceedings of the IEEE/CVF Conference on Computer Vision and Pattern Recognition Workshops}, pages 996--997, 2020.

\bibitem[Zhou and Tuzel(2018)]{zhou2018voxelnet}
Yin Zhou and Oncel Tuzel.
\newblock Voxelnet: End-to-end learning for point cloud based 3d object detection.
\newblock In \emph{Proceedings of the IEEE conference on computer vision and pattern recognition}, pages 4490--4499, 2018.

\bibitem[Lang et~al.(2019)Lang, Vora, Caesar, Zhou, Yang, and Beijbom]{lang2019pointpillars}
Alex~H Lang, Sourabh Vora, Holger Caesar, Lubing Zhou, Jiong Yang, and Oscar Beijbom.
\newblock Pointpillars: Fast encoders for object detection from point clouds.
\newblock In \emph{Proceedings of the IEEE/CVF conference on computer vision and pattern recognition}, pages 12697--12705, 2019.

\bibitem[Philion and Fidler(2020)]{philion2020lift}
Jonah Philion and Sanja Fidler.
\newblock Lift, splat, shoot: Encoding images from arbitrary camera rigs by implicitly unprojecting to 3d.
\newblock In \emph{Computer Vision--ECCV 2020: 16th European Conference, Glasgow, UK, August 23--28, 2020, Proceedings, Part XIV 16}, pages 194--210. Springer, 2020.

\bibitem[Liu et~al.(2022{\natexlab{a}})Liu, Wang, Zhang, and Sun]{liu2022petr}
Yingfei Liu, Tiancai Wang, Xiangyu Zhang, and Jian Sun.
\newblock Petr: Position embedding transformation for multi-view 3d object detection.
\newblock In \emph{European Conference on Computer Vision}, pages 531--548. Springer, 2022{\natexlab{a}}.

\bibitem[Vaswani(2017)]{vaswani2017attention}
A~Vaswani.
\newblock Attention is all you need.
\newblock \emph{Advances in Neural Information Processing Systems}, 2017.

\bibitem[Park et~al.(2022)Park, Xu, Yang, Keutzer, Kitani, Tomizuka, and Zhan]{park2022time}
Jinhyung Park, Chenfeng Xu, Shijia Yang, Kurt Keutzer, Kris~M Kitani, Masayoshi Tomizuka, and Wei Zhan.
\newblock Time will tell: New outlooks and a baseline for temporal multi-view 3d object detection.
\newblock In \emph{The Eleventh International Conference on Learning Representations}, 2022.

\bibitem[Yang et~al.(2023)Yang, Chen, Tian, Tao, Zhu, Zhang, Huang, Li, Qiao, Lu, et~al.]{yang2023bevformer}
Chenyu Yang, Yuntao Chen, Hao Tian, Chenxin Tao, Xizhou Zhu, Zhaoxiang Zhang, Gao Huang, Hongyang Li, Yu~Qiao, Lewei Lu, et~al.
\newblock Bevformer v2: Adapting modern image backbones to bird's-eye-view recognition via perspective supervision.
\newblock In \emph{Proceedings of the IEEE/CVF Conference on Computer Vision and Pattern Recognition}, pages 17830--17839, 2023.

\bibitem[Hong et~al.(2022)Hong, Dai, and Ding]{yuhong-CMKD-ECCV2022}
Yu~Hong, Hang Dai, and Yong Ding.
\newblock Cross-modality knowledge distillation network for monocular 3d object detection.
\newblock In \emph{{ECCV}}, Lecture Notes in Computer Science. Springer, 2022.

\bibitem[Huang et~al.(2022)Huang, Liu, Zhang, Zhang, Xu, Wang, and Liu]{huang2022tig}
Peixiang Huang, Li~Liu, Renrui Zhang, Song Zhang, Xinli Xu, Baichao Wang, and Guoyi Liu.
\newblock Tig-bev: Multi-view bev 3d object detection via target inner-geometry learning.
\newblock \emph{arXiv preprint arXiv:2212.13979}, 2022.

\bibitem[Chen et~al.(2022)Chen, Li, Zhang, Fang, Jiang, and Zhao]{chen2022bevdistill}
Zehui Chen, Zhenyu Li, Shiquan Zhang, Liangji Fang, Qinhong Jiang, and Feng Zhao.
\newblock Bevdistill: Cross-modal bev distillation for multi-view 3d object detection.
\newblock \emph{arXiv preprint arXiv:2211.09386}, 2022.

\bibitem[Jang et~al.(2024)Jang, Jo, Hwang, Lee, and Ji]{jang2024stxd}
Sujin Jang, Dae~Ung Jo, Sung~Ju Hwang, Dongwook Lee, and Daehyun Ji.
\newblock Stxd: Structural and temporal cross-modal distillation for multi-view 3d object detection.
\newblock \emph{Advances in Neural Information Processing Systems}, 36, 2024.

\bibitem[Wang et~al.(2020)Wang, Chen, You, Li, Hariharan, Campbell, Weinberger, and Chao]{wang2020train}
Yan Wang, Xiangyu Chen, Yurong You, Li~Erran Li, Bharath Hariharan, Mark Campbell, Kilian~Q Weinberger, and Wei-Lun Chao.
\newblock Train in germany, test in the usa: Making 3d object detectors generalize.
\newblock In \emph{Proceedings of the IEEE/CVF Conference on Computer Vision and Pattern Recognition}, pages 11713--11723, 2020.

\bibitem[Yang et~al.(2021)Yang, Shi, Wang, Li, and Qi]{yang2021st3d}
Jihan Yang, Shaoshuai Shi, Zhe Wang, Hongsheng Li, and Xiaojuan Qi.
\newblock St3d: Self-training for unsupervised domain adaptation on 3d object detection.
\newblock In \emph{Proceedings of the IEEE/CVF conference on computer vision and pattern recognition}, pages 10368--10378, 2021.

\bibitem[Xu et~al.(2021)Xu, Zhou, Wang, Qi, and Anguelov]{xu2021spg}
Qiangeng Xu, Yin Zhou, Weiyue Wang, Charles~R Qi, and Dragomir Anguelov.
\newblock Spg: Unsupervised domain adaptation for 3d object detection via semantic point generation.
\newblock In \emph{Proceedings of the IEEE/CVF International Conference on Computer Vision}, pages 15446--15456, 2021.

\bibitem[Yihan et~al.(2021)Yihan, Wang, Wang, Xu, Ye, Yang, and Ma]{yihan2021learning}
Zeng Yihan, Chunwei Wang, Yunbo Wang, Hang Xu, Chaoqiang Ye, Zhen Yang, and Chao Ma.
\newblock Learning transferable features for point cloud detection via 3d contrastive co-training.
\newblock \emph{Advances in Neural Information Processing Systems}, 34:\penalty0 21493--21504, 2021.

\bibitem[Wei et~al.(2022{\natexlab{a}})Wei, Wei, Rao, Li, Zhou, and Lu]{wei2022lidar}
Yi~Wei, Zibu Wei, Yongming Rao, Jiaxin Li, Jie Zhou, and Jiwen Lu.
\newblock Lidar distillation: Bridging the beam-induced domain gap for 3d object detection.
\newblock In \emph{European Conference on Computer Vision}, pages 179--195. Springer, 2022{\natexlab{a}}.

\bibitem[Zhang et~al.(2024)Zhang, Zhou, and Huang]{zhang2024stal3d}
Yanan Zhang, Chao Zhou, and Di~Huang.
\newblock Stal3d: Unsupervised domain adaptation for 3d object detection via collaborating self-training and adversarial learning.
\newblock \emph{IEEE Transactions on Intelligent Vehicles}, 2024.

\bibitem[Hu et~al.(2023)Hu, Liu, and Hu]{hu2023density}
Qianjiang Hu, Daizong Liu, and Wei Hu.
\newblock Density-insensitive unsupervised domain adaption on 3d object detection.
\newblock In \emph{Proceedings of the IEEE/CVF Conference on Computer Vision and Pattern Recognition}, pages 17556--17566, 2023.

\bibitem[Vidit et~al.(2023)Vidit, Engilberge, and Salzmann]{vidit2023clip}
Vidit Vidit, Martin Engilberge, and Mathieu Salzmann.
\newblock Clip the gap: A single domain generalization approach for object detection.
\newblock In \emph{Proceedings of the IEEE/CVF Conference on Computer Vision and Pattern Recognition}, pages 3219--3229, 2023.

\bibitem[Wu and Deng(2022)]{wu2022single}
Aming Wu and Cheng Deng.
\newblock Single-domain generalized object detection in urban scene via cyclic-disentangled self-distillation.
\newblock In \emph{Proceedings of the IEEE/CVF Conference on computer vision and pattern recognition}, pages 847--856, 2022.

\bibitem[He et~al.(2022)He, Chen, Xie, Li, Doll{\'a}r, and Girshick]{he2022masked}
Kaiming He, Xinlei Chen, Saining Xie, Yanghao Li, Piotr Doll{\'a}r, and Ross Girshick.
\newblock Masked autoencoders are scalable vision learners.
\newblock In \emph{Proceedings of the IEEE/CVF conference on computer vision and pattern recognition}, pages 16000--16009, 2022.

\bibitem[Chen et~al.(2020)Chen, Kornblith, Norouzi, and Hinton]{chen2020simple}
Ting Chen, Simon Kornblith, Mohammad Norouzi, and Geoffrey Hinton.
\newblock A simple framework for contrastive learning of visual representations.
\newblock In \emph{International conference on machine learning}, pages 1597--1607. PMLR, 2020.

\bibitem[Li et~al.(2022{\natexlab{b}})Li, Chen, Li, Fang, Jiang, Liu, and Jiang]{li2022unsupervised}
Zhenyu Li, Zehui Chen, Ang Li, Liangji Fang, Qinhong Jiang, Xianming Liu, and Junjun Jiang.
\newblock Unsupervised domain adaptation for monocular 3d object detection via self-training.
\newblock In \emph{European conference on computer vision}, pages 245--262. Springer, 2022{\natexlab{b}}.

\bibitem[Houlsby et~al.(2019)Houlsby, Giurgiu, Jastrzebski, Morrone, De~Laroussilhe, Gesmundo, Attariyan, and Gelly]{houlsby2019parameter}
Neil Houlsby, Andrei Giurgiu, Stanislaw Jastrzebski, Bruna Morrone, Quentin De~Laroussilhe, Andrea Gesmundo, Mona Attariyan, and Sylvain Gelly.
\newblock Parameter-efficient transfer learning for nlp.
\newblock In \emph{International conference on machine learning}, pages 2790--2799. PMLR, 2019.

\bibitem[Lian et~al.(2022)Lian, Zhou, Feng, and Wang]{lian2022scaling}
Dongze Lian, Daquan Zhou, Jiashi Feng, and Xinchao Wang.
\newblock Scaling \& shifting your features: A new baseline for efficient model tuning.
\newblock \emph{Advances in Neural Information Processing Systems}, 35:\penalty0 109--123, 2022.

\bibitem[Liu et~al.(2022{\natexlab{b}})Liu, Tam, Muqeeth, Mohta, Huang, Bansal, and Raffel]{liu2022few}
Haokun Liu, Derek Tam, Mohammed Muqeeth, Jay Mohta, Tenghao Huang, Mohit Bansal, and Colin~A Raffel.
\newblock Few-shot parameter-efficient fine-tuning is better and cheaper than in-context learning.
\newblock \emph{Advances in Neural Information Processing Systems}, 35:\penalty0 1950--1965, 2022{\natexlab{b}}.

\bibitem[Dong et~al.(2024)Dong, Guo, Liu, Zhang, and Zhang]{dong2024ppea}
Yue-Jiang Dong, Yuan-Chen Guo, Ying-Tian Liu, Fang-Lue Zhang, and Song-Hai Zhang.
\newblock Ppea-depth: Progressive parameter-efficient adaptation for self-supervised monocular depth estimation.
\newblock In \emph{Proceedings of the AAAI Conference on Artificial Intelligence}, volume~38, pages 1609--1617, 2024.

\bibitem[Dosovitskiy et~al.(2017)Dosovitskiy, Ros, Codevilla, Lopez, and Koltun]{Dosovitskiy17}
Alexey Dosovitskiy, German Ros, Felipe Codevilla, Antonio Lopez, and Vladlen Koltun.
\newblock {CARLA}: {An} open urban driving simulator.
\newblock In \emph{Proceedings of the 1st Annual Conference on Robot Learning}, pages 1--16, 2017.

\bibitem[Gu et~al.(2021)Gu, Zhou, Xu, Feng, Cheng, Lu, Shi, and Ma]{gu2021pit}
Qiqi Gu, Qianyu Zhou, Minghao Xu, Zhengyang Feng, Guangliang Cheng, Xuequan Lu, Jianping Shi, and Lizhuang Ma.
\newblock Pit: Position-invariant transform for cross-fov domain adaptation.
\newblock In \emph{Proceedings of the IEEE/CVF International Conference on Computer Vision}, pages 8761--8770, 2021.

\bibitem[Klinghoffer et~al.(2023)Klinghoffer, Philion, Chen, Litany, Gojcic, Joo, Raskar, Fidler, and Alvarez]{klinghoffer2023towards}
Tzofi Klinghoffer, Jonah Philion, Wenzheng Chen, Or~Litany, Zan Gojcic, Jungseock Joo, Ramesh Raskar, Sanja Fidler, and Jose~M Alvarez.
\newblock Towards viewpoint robustness in bird's eye view segmentation.
\newblock In \emph{Proceedings of the IEEE/CVF International Conference on Computer Vision}, pages 8515--8524, 2023.

\bibitem[Zhao et~al.(2021)Zhao, Kong, and Fowlkes]{zhao2021camera}
Yunhan Zhao, Shu Kong, and Charless Fowlkes.
\newblock Camera pose matters: Improving depth prediction by mitigating pose distribution bias.
\newblock In \emph{Proceedings of the IEEE/CVF Conference on Computer Vision and Pattern Recognition}, pages 15759--15768, 2021.

\bibitem[Wang et~al.(2019)Wang, Fang, Qian, Yang, Zhou, and Zhou]{wang2019perspective}
Ke~Wang, Bin Fang, Jiye Qian, Su~Yang, Xin Zhou, and Jie Zhou.
\newblock Perspective transformation data augmentation for object detection.
\newblock \emph{IEEE Access}, 8:\penalty0 4935--4943, 2019.

\bibitem[Godard et~al.(2019)Godard, Mac~Aodha, Firman, and Brostow]{godard2019digging}
Cl{\'e}ment Godard, Oisin Mac~Aodha, Michael Firman, and Gabriel~J Brostow.
\newblock Digging into self-supervised monocular depth estimation.
\newblock In \emph{Proceedings of the IEEE/CVF international conference on computer vision}, pages 3828--3838, 2019.

\bibitem[Bian et~al.(2023)Bian, Wang, Li, Bian, and Prisacariu]{bian2023nope}
Wenjing Bian, Zirui Wang, Kejie Li, Jia-Wang Bian, and Victor~Adrian Prisacariu.
\newblock Nope-nerf: Optimising neural radiance field with no pose prior.
\newblock In \emph{Proceedings of the IEEE/CVF Conference on Computer Vision and Pattern Recognition}, pages 4160--4169, 2023.

\bibitem[Lyu et~al.(2021)Lyu, Liu, Wang, Kong, Liu, Liu, Chen, and Yuan]{lyu2021hr}
Xiaoyang Lyu, Liang Liu, Mengmeng Wang, Xin Kong, Lina Liu, Yong Liu, Xinxin Chen, and Yi~Yuan.
\newblock Hr-depth: High resolution self-supervised monocular depth estimation.
\newblock In \emph{Proceedings of the AAAI conference on artificial intelligence}, volume~35, pages 2294--2301, 2021.

\bibitem[Zhou et~al.(2021)Zhou, Greenwood, and Taylor]{zhou2021self}
Hang Zhou, David Greenwood, and Sarah Taylor.
\newblock Self-supervised monocular depth estimation with internal feature fusion.
\newblock \emph{arXiv preprint arXiv:2110.09482}, 2021.

\bibitem[Wei et~al.(2022{\natexlab{b}})Wei, Zhao, Zheng, Zhu, Rao, Huang, Lu, and Zhou]{wei2022surround}
Yi~Wei, Linqing Zhao, Wenzhao Zheng, Zheng Zhu, Yongming Rao, Guan Huang, Jiwen Lu, and Jie Zhou.
\newblock Surrounddepth: Entangling surrounding views for self-supervised multi-camera depth estimation.
\newblock \emph{arXiv preprint arXiv:2204.03636}, 2022{\natexlab{b}}.

\bibitem[Wang et~al.(2004)Wang, Bovik, Sheikh, and Simoncelli]{wang2004image}
Zhou Wang, Alan~C Bovik, Hamid~R Sheikh, and Eero~P Simoncelli.
\newblock Image quality assessment: from error visibility to structural similarity.
\newblock \emph{IEEE transactions on image processing}, 13\penalty0 (4):\penalty0 600--612, 2004.

\bibitem[He et~al.(2016)He, Zhang, Ren, and Sun]{he2016deep}
Kaiming He, Xiangyu Zhang, Shaoqing Ren, and Jian Sun.
\newblock Deep residual learning for image recognition.
\newblock In \emph{Proceedings of the IEEE conference on computer vision and pattern recognition}, pages 770--778, 2016.

\bibitem[Zhu et~al.(2019)Zhu, Jiang, Zhou, Li, and Yu]{zhu2019class}
Benjin Zhu, Zhengkai Jiang, Xiangxin Zhou, Zeming Li, and Gang Yu.
\newblock Class-balanced grouping and sampling for point cloud 3d object detection.
\newblock 2019.

\end{thebibliography}

\clearpage
\appendix

\begin{appendices}

\vspace{-4em}
\part{} 
\parttoc 

\section{Datasets}
\label{apx:dataset}

\begin{table}[h]
    \caption{Dataset details. Note that each statistical information is calculated from the whole dataset.}
    \label{tab:dataset}
    \resizebox{\linewidth}{!}{%
    \begin{tabular}{l|c|c|c|c|c|c|c|c|c|c} 
    \toprule
        Dataset  & Cameras & LiDAR & $\#$ scenes & $\#$ 3D boxes & Points per Beam & Range & Location & Night & Rain & Highway \\\hline
        nuScenes  & 6 & 32-beam & 1000 & 1.4M & 1,084 & $<100\text{m}$ & USA and Singapore & $\checkmark$ & $\checkmark$ & - \\\hline
        Lyft & 6 & 64-beam & 366 & 1.3M & 1,863 & $<100\text{m}$ & USA & - & $\checkmark$ & - \\\hline
        Waymo & 5 & 64-beam & 1150 & 12M & 2,258 & $<100\text{m}$ & USA & $\checkmark$ & $\checkmark$ & - \\\hline
        CARLA & 6 & 128-beam & 10 & 2.0M & 2,500 & $<100\text{m}$ & Carla Town10 & - & - & - \\
       \bottomrule
    \end{tabular}}
\end{table}

We evaluate overall performance on landmark datasets for 3D Object Detection: Waymo~\cite{sun2020scalability}, Lyft~\cite{houston2021one}, and nuScenes~\cite{caesar2020nuscenes}. The three datasets have different point cloud ranges and specifications. Hence, we convert them to a unified range and coordinates for accurate comparison. We also adopt only seven parameters to achieve consistent training results under the same conditions: center locations $(x, y, z)$, box size $(l, w, h)$, and heading angle $\theta$. Additionally, to estimate practical degradation due to changes in camera positioning, we conducted a proof of concept by generating data similar to the nuScenes using the CARLA simulation. The details are as follows:

\noindent\textbf{Waymo}
The Waymo dataset~\cite{sun2020scalability} consists of high-quality and large-scale data with $230$K frames from all 1,150 scenes using multiple LiDAR scanners and cameras. Furthermore, for the generalization purpose, Waymo is recorded at diverse cities, weather conditions, and times. For object detection in 2D or 3D, Waymo provides point cloud-annotated 3D bounding boxes as 3D data pairs and RGB image-annotated 2D bounding boxes as 2D data pairs. 

\noindent\textbf{nuScenes}
The nuScenes dataset~\cite{caesar2020nuscenes} uses 6 cameras that cover a full 360-degree range of view and a single LiDAR sensor to obtain $40$K frames from 20-second-long 1,000 video sequences, which are fully annotated with 3D bounding boxes for 10 object classes. The nuScenes dataset covers 28k annotated samples for training. Also, validation and test contain 6k scenes each. The nuScenes frames are captured in the same manner as Waymo dataset for the data diversity. But unlike Waymo, nuScenes provides labels only for the point cloud data with 23 classes of 3D bounding boxes.

\noindent\textbf{Lyft}
The Lyft dataset~\cite{houston2021one} is motivated by the impact of large-scale datasets on Machine Learning and consists of over 1,000 hours of data. This was collected by a fleet of 20 autonomous vehicles along a fixed route in Palo Alto, California, over a four-month period. It consists of 170,000 scenes (each scene is 25 seconds long) and contains 3D bounding boxes with the precise positions of nearby vehicles, cyclists, and pedestrians over time. In addition, the Lyft dataset includes a high-definition semantic map with 15,242 labelled elements and a high-definition aerial view over the area. 

\begin{figure}
    \centering
    \includegraphics[width=0.8\linewidth]{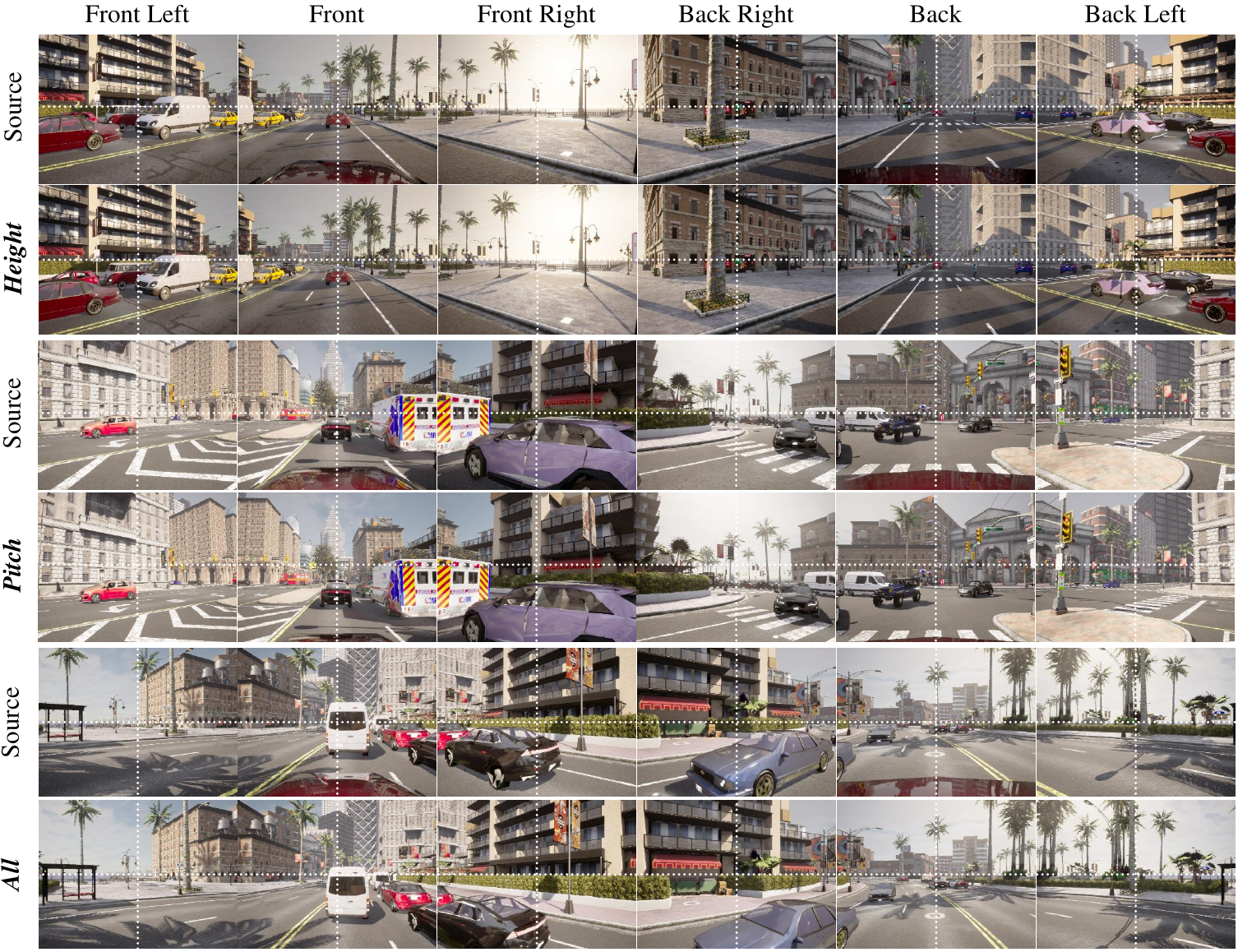}
    \caption{The paired sample of each evaluation set in Carla dataset.}
    \label{fig:carla_appendix}
\end{figure}

\noindent\textbf{CARLA}
To quantify the performance drop resulting from camera shifts, we employed an autonomous driving simulation powered by CARLA~\cite{Dosovitskiy17} 0.9.14 and Unreal Engine 4.26.
We collected 24K frames for training and 1K frames for each evaluation, driving through Town10 under cloudless weather conditions between sunrise and sunset times. This dataset includes over 100 vehicles and 30 pedestrians in random locations. In Fig.~\ref{fig:carla_appendix}, the Source utilizes 6 nuScenes-like cameras and 6 LiDARs, while the \textit{Target} has perturbed sensors. From the Source sensors, the \textit{Height} increases by 0.65m and the
\textit{Pitch} increases by 5 degrees. The \textit{All} synthetically moves the x, y, z-coordinates by -0.12m, 0.65m, and -0.2m/+0.2m, respectively, and rotates the yaw by -5/+5 degrees, depending on their directions.
Each target sets is collected simultaneously with the Source.

\section{Implementation Details}
\label{apx:implementation}
To validate the effectiveness of our proposed methods, we adopt BEVDepth~\cite{li2023bevdepth} and BEVFormer~\cite{li2022bevformer} as our base detectors.
Both detectors utilize ResNet50~\cite{he2016deep} backbone that initialized from ImageNet-1K.
Also, we construct BEV representations within a perception range of [-50.0m, 50.0m] for both the X and Y axes.
In BEVDepth, we reshape multi-view input image resolutions as follow: $[256, 704]$ for nuScenes, $[384, 704]$ for Lyft, $[320, 704]$ for Waymo.
As following DG-BEV~\cite{wang2023towards}, we train 24 epochs with AdamW optimizer by learning rate 2e-4 in pre-training phase. The training takes approximately 18 hours using one A100 GPU.
In fine-tuning phase, we conduct an extensive grid search to determine the optimal learning rate proportional to the number of learnable parameters.
Note that we extensively augment various image conditions as detailed in~\cite{wang2023towards}.

\section{Additional Experiments}
\label{apx:add_exp}

In this appendix, we present additional experiments to validate the effectiveness of our proposed paradigm.
First, Tab.~\ref{tab:UDGA2} summarizes the overall results of our work from the perspective of domain shift. We also analyze how changes in camera positioning worsen the performance and evaluate whether existing augmentation methods can mitigate the deterioration. Additionally, we conduct ablation studies to enhance the LEDA structure, including comparisons with formal adapters.
Finally, we present the comparison results with the transformer-based detector. The qualitative analysis of the multi-view results from our proposed paradigm is included towards the end of this chapter.


\begin{table}[t]
    \caption{Comparison of Unified Domain Generalization and Adaptation performance with state-of-the-art techniques. We validate our proposed methods with the same baseline model, named BEVDepth, on Cross-domain. The \textbf{bold} values indicate the best performance. Also, --- denotes \textit{`Do not support'}.}
    \label{tab:UDGA2}
    \centering
    \resizebox{\linewidth}{!}{%
    \begin{tabular}{p{3.5cm}p{3cm}p{2.5cm}|>{\centering\arraybackslash}p{3cm}|>{\centering\arraybackslash}p{3cm}}
        \toprule 
        \multirow{2.3}{*}{Task} & \multirow{2.3}{*}{Method} & \multirow{2.3}{*}{Branch} & Source & Target \\ 
        \cmidrule{4-5}
        & & & ~NDS$^{\hat{*}}$$\uparrow$ / mAP$\uparrow$~ & ~NDS$^{\hat{*}}$$\uparrow$ / mAP$\uparrow$~ \\
        \midrule
        \multirow{8}{*}{Lyft $\rightarrow$ nuScenes}
        & \textit{Direct Transfer}&   & 0.684 / 0.602 & 0.213 / 0.102 \\
        & \textit{Oracle}         &   & 0.296 / 0.112 & 0.587 / 0.475 \\
        \cmidrule{2-5}
        & DG-BEV~\cite{wang2023towards}                                         & DG & 0.675 / 0.611 & 0.374 / 0.268 \\
        & PD-BEV~\cite{lu2023towards}                                        & DG & 0.677 / 0.593 & 0.344 / 0.263 \\
        & PD-BEV                                         & UDA & 0.672 / 0.589 & 0.358 / 0.280 \\
        \cmidrule{2-5}
        \coloredrowcell{DCDCDC} & Ours                    & DG & \textbf{0.702} / \textbf{0.630} & 0.421 / 0.281 \\
        \coloredrowcell{DCDCDC} & Ours (1$\%$)             & UDGA & \textbf{0.702} / \textbf{0.630} & 0.526 / 0.404 \\
        \coloredrowcell{DCDCDC} & Ours (5$\%$)            & UDGA & \textbf{0.702} / \textbf{0.630} & \textbf{0.563} / \textbf{0.444} \\
        \midrule
        \multirow{8}{*}{nuScenes $\rightarrow$ Lyft} 
        & \textit{Direct Transfer}&             & 0.587 / 0.475 & 0.296 / 0.112 \\
        & \textit{Oracle}         &             & 0.213 / 0.102 & 0.684 / 0.602 \\
        \cmidrule{2-5}
        & DG-BEV                                         & DG          & 0.578 / 0.470 & 0.437 / 0.287 \\
        & PD-BEV                                         & DG          & --- & 0.458 / 0.304 \\
        & PD-BEV                                         & UDA         & --- & 0.476 / 0.316 \\        
        \cmidrule{2-5}
        \coloredrowcell{DCDCDC} & Ours                    & DG          & \textbf{0.623} / \textbf{0.513} 
        & 0.487 / 0.324 \\
        \coloredrowcell{DCDCDC} & Ours (1$\%$)            & UDGA & \textbf{0.623} / \textbf{0.513} & 0.578 / 0.462 \\
        \coloredrowcell{DCDCDC} & Ours (5$\%$)            & UDGA & \textbf{0.623} / \textbf{0.513} & \textbf{0.613} / \textbf{0.506} \\
        \midrule
        \multirow{7}{*}{Waymo $\rightarrow$ nuScenes} 
        & \textit{Direct Transfer}&   & 0.649 / 0.552 & 0.133 / 0.032 \\
        & \textit{Oracle}         &   & 0.178 / 0.040 & 0.587 / 0.475 \\
        \cmidrule{2-5}
        & DG-BEV                                         & DG & \textbf{0.660} / \textbf{0.568} & 0.472 / 0.303 \\
        \cmidrule{2-5}
        \coloredrowcell{DCDCDC} & Ours                    & DG & 0.656 / 0.547 & 0.477 / 0.326 \\
        \coloredrowcell{DCDCDC} & Ours (1$\%$)            & UDGA & 0.656 / 0.547 & 0.534 / 0.409 \\
        \coloredrowcell{DCDCDC} & Ours (5$\%$)            & UDGA & 0.656 / 0.547 & \textbf{0.571} / \textbf{0.448} \\
        \midrule
        \multirow{7}{*}{nuScenes $\rightarrow$ Waymo} 
        & \textit{Direct Transfer}&   & 0.587 / 0.475 & 0.178 / 0.040 \\
        & \textit{Oracle}         &   & 0.133 / 0.032 & 0.649 / 0.552 \\
        \cmidrule{2-5}
        & DG-BEV                                         & DG & 0.563 / 0.461 & 0.415 / 0.297 \\
        \cmidrule{2-5}
        \coloredrowcell{DCDCDC} & Ours                    & DG & \textbf{0.603} / \textbf{0.497} & 0.459 / 0.349 \\
        \coloredrowcell{DCDCDC} & Ours (1$\%$)            & UDGA & \textbf{0.603} / \textbf{0.497} & 0.509 / 0.378 \\
        \coloredrowcell{DCDCDC} & Ours (5$\%$)            & UDGA & \textbf{0.603} / \textbf{0.497} & \textbf{0.549} / \textbf{0.424} \\
        \bottomrule
    \end{tabular}}
\end{table}

\myparagraph{Performance across domains.}
In this section, we compare our proposed UDGA with existing solutions (\ie, DG, UDA) in various cross-domain conditions (see Tab.~\ref{tab:UDGA2}).
We aim to practically mitigate perspective shifts without hindering well-defined source knowledge. Our DG branch achieves top performance, surpassing \textit{Direct Transfer} in the Source domain. The UDGA, which follows DG, improves Target accuracy without compromising Source performance.
Especially, we advocate that UDGA enables efficient adaptation with significantly down-scaled data split (\ie, 1$\%$ and 5$\%$).
Also, it is noteworthy that UDGA do not forget previously learned potentials, fully transferring to target domains (up to +14.2$\%$ NDS gain in Lyft$\rightarrow$nuScenes). 
Overall, UDGA provide a practical solution to address perspective view changes, efficiently adapting with only tiny split.


\begin{table}[h]
    \caption{Performance under CALRA-simulated domain changes. The model is trained exclusively on Source.
    The \textit{diff} shows the Source-Target difference. The \textbf{bold} values indicate the worst difference.}
    \label{tab:carla_exp}
    \centering
    \resizebox{0.6\linewidth}{!}{%
        \begin{tabular}{p{2cm}|ccccc}
            \toprule
             Test domain & NDS$^{\hat{*}}$$\uparrow$ & mAP$\uparrow$& mATE$\downarrow$ & mASE$\downarrow$ & mAOE$\downarrow$ \\
            \cmidrule{1-6}
            Source & 0.666 & 0.811 & 0.229 & 0.122 & 0.043 \\
            Target:\textit{Pitch}  & 0.449 & 0.491 & 0.739 & 0.159 & 0.065 \\            
            \rowcolor[HTML]{DCDCDC}\textit{diff} & -0.217 & -0.319 & 0.510 & 0.036 & 0.022 \\
            \cmidrule{1-6}
            Source & 0.688 & 0.848 & 0.210 & 0.111 & 0.042 \\
            Target:\textit{Height} & 0.313 & 0.280 & 1.362 & 0.179 & 0.090 \\
            \rowcolor[HTML]{DCDCDC}\textit{diff} & \textbf{-0.374} & \textbf{-0.568} & \textbf{1.152} & 0.067 & 0.048 \\
            \cmidrule{1-6}
            Source     & 0.687 & 0.847 & 0.211 & 0.216 & 0.372 \\
            Target:\textit{All} & 0.321 & 0.301 & 1.357 & 0.181 & 0.110 \\
            \rowcolor[HTML]{DCDCDC}\textit{diff} & -0.366 & -0.546 & 1.146 & \textbf{0.069} & \textbf{0.071} \\
           \bottomrule
        \end{tabular}}
\end{table}
\begin{table}[h!]
    \caption{Performance of multi-view augmentations in domain shift. Gray highlight denotes `Ours'.}
    \label{tab:augmentation}
    \centering
    \resizebox{0.6\linewidth}{!}{%
    \begin{tabular}{l|cc|cc}
        \toprule 
        \multirow{2.3}{*}{Method} & \multicolumn{2}{c|}{Lyft $\rightarrow$ nuScenes} & \multicolumn{2}{c}{nuScenes $\rightarrow$ Lyft} \\
        \cmidrule{2-5}
        & ~NDS$^{\hat{*}}$$\uparrow$~ & ~mAP$\uparrow$~ & ~NDS$^{\hat{*}}$$\uparrow$~ & ~mAP$\uparrow$~ \\
        \midrule
        \textit{Direct Transfer}                & 0.213 & 0.102 & 0.296 & 0.112 \\
        GT sampling                             & 0.269 & 0.211 & 0.405 & 0.263 \\
        2D augmentation                                 & 0.269 & 0.221 & 0.423 & 0.263 \\
        3D augmentation                                 & 0.289 & 0.235 & 0.403 & 0.243 \\
        Extrinsic augmentation                          & 0.298 & 0.223 & 0.436 & 0.255 \\
        CBGS~\cite{zhu2019class}                                    & 0.265 & 0.196 & 0.349 & 0.215 \\
        \cmidrule{1-5}
        DG-BEV                                  & 0.374 & 0.268 & 0.437 & 0.287 \\
        \rowcolor[HTML]{DCDCDC}Ours             & \textbf{0.421} & \textbf{0.281} & \textbf{0.487} & \textbf{0.324} \\        
        \bottomrule
    \end{tabular}}
\end{table}

\myparagraph{Practical domain shift analysis.}
We analyze the impact of changes in camera geometry on 3D object estimation. The experimental model is trained using only the source dataset on ResNet50-based BEVDet and then evaluated on three sets of (source, target) to analyze performance differences. In Tab.~\ref{tab:carla_exp}, the performance of the source is similar in all test sets. On the other hand, the performance of the target decreases significantly in all cases. Since this experiment is conducted in the same environment with the same camera sensors, it demonstrates how much performance degradation is caused by the position of the camera. The set with the largest performance drop in the target is \textit{Height}, where the mATE value increased significantly. The target \textit{All} exhibits the worst mASE and mAOE, while the other measures also deteriorated by a similar amount as \textit{Height}.

Conventional augmentation methods enhance the robustness of the model. We evaluate some of them in Tab.~\ref{tab:augmentation}. GT sampling and CBGS~\cite{zhu2019class} are techniques designed to balance ground truths. 2D augmentation directly augment multi-view inputs (i.e., image resize, crop and paste, contrast and brightness distortion).
3D and extrinsic methods are global augmentations that address both input and ground truths, and ground truths only, respectively. 
These methods enhance geometric understanding from input noises. However, in dynamic view changes (i.e., cross-domain), they still suffer from geometric inconsistency and show poor generalization capability. Moreover, various 2D approaches do not guarantee geometric alignments between 2D images and 3D ground truths and relevant studies have not been explored well, as reported in~\cite{wang2023towards} and~\cite{zhao2021camera}. 

\begin{table}[h!]
    \caption{Performance comparison for each module (UDGA 5$\%$). Gray highlight denotes `Ours'.}
    \label{tab:da_module}
    \centering
    \resizebox{\linewidth}{!}{%
    \begin{tabular}{c|c|c|c|cc|cc}
        \toprule 
        \multirow{2.3}{*}{Backbone} & \multirow{2.3}{*}{View transform} & \multirow{2.3}{*}{BEV encoder} & \multirow{2.3}{*}{Detection head}& \multicolumn{2}{c|}{Lyft $\rightarrow$ nuScenes} & \multicolumn{2}{c}{nuScenes $\rightarrow$ Lyft} \\ 
        \cmidrule{5-8}
        &&&& ~NDS$^{\hat{*}}$$\uparrow$~ & ~mAP$\uparrow$~ & ~NDS$^{\hat{*}}$$\uparrow$~ & ~mAP$\uparrow$~ \\
        \midrule
                    &             &             &            & 0.421 & 0.281 & 0.487 & 0.324 \\
                    &             &             & \checkmark & 0.333 & 0.237 & 0.489 & 0.352 \\
                    &             & \checkmark  & \checkmark & 0.433 & 0.322 & 0.551 & 0.418 \\
                    & \checkmark  & \checkmark  & \checkmark & 0.525 & 0.409 & 0.608 & 0.498 \\
        \rowcolor[HTML]{DCDCDC} \checkmark & \checkmark  & \checkmark  & \checkmark & \textbf{0.563} & \textbf{0.444} & \textbf{0.613} & \textbf{0.506} \\
        \bottomrule
    \end{tabular}}
\end{table}
\begin{table}[h!]
    \caption{Comparison with various adapter structures (UDGA 10$\%$). Gray highlight denotes `Ours'.}
    \label{tab:da_structure}
    \centering
    \resizebox{0.9\linewidth}{!}{%
    \begin{tabular}{c|c|c|c|cc|cc}
        \toprule 
        \multirow{2.3}{*}{Method} &  \multirow{2.3}{*}{Project Down} &  \multirow{2.3}{*}{Project Up} &  \multirow{2.3}{*}{$\#$ Params} & \multicolumn{2}{c|}{Lyft $\rightarrow$ nuScenes} & \multicolumn{2}{c}{nuScenes $\rightarrow$ Lyft} \\
        \cmidrule{5-8}
        &&&& ~NDS$^{\hat{*}}$$\uparrow$~ & ~mAP$\uparrow$~ & ~NDS$^{\hat{*}}$$\uparrow$~ & ~mAP$\uparrow$~ \\
        \midrule
        Adapter-H & Conv.  & Conv.  & 25.9M & 0.547 & 0.439 & 0.592 & 0.484 \\
        Adapter-B & Conv.  & Linear & 21.3M & 0.511 & 0.384 & 0.584 & 0.475 \\
        Adapter-S & Linear & Conv.  & 8.8M  & 0.444 & 0.255 & 0.500 & 0.356 \\
        Adapter-T & Linear & Linear & 2.9M  & 0.282 & 0.262 & 0.398 & 0.376 \\
        \rowcolor[HTML]{DCDCDC} Ours & Conv. & Linear & 8.8M & \textbf{0.573} & \textbf{0.457} & \textbf{0.638} & \textbf{0.537} \\
        \bottomrule
    \end{tabular}}
\end{table}

\myparagraph{Searching adapter structures.}
We explore various modules and structures to find a suitable adapter architecture. Tab.~\ref{tab:da_module},~\ref{tab:da_structure} show which structures and locations affects the model's performance. For adapter locations, performance is optimal when adapters are attached to all modules, gradually improving with the addition of more. Exceptionally, attaching adapters only at the Detection Head leads to a decline in Lyft$\rightarrow$nuScenes. In addition, Tab.~\ref{tab:da_structure} represents the performance of various adapter structures. The combination of Convolution and Linear layer respectively for Project Down and Up shows the best performance in both tasks. Note that training with fewer parameters(8.8M) is more effective. However, we suggest that large-scale parameters may require a larger dataset or more training, as we only trained on 10$\%$ of the target dataset for less than 20 epochs in this experiment.


\begin{table}[h]
    \caption{Comparison of UDGA performance on BEVFormer. We train with two different data splits 50$\%$, and 100$\%$. Additionally, $\#$ Params denote the number of parameters for training. The bold values indicate the best performance. --- denotes \textit{`Do not support'}.}
    \label{tab:DA2}
    \centering
    \resizebox{0.7\linewidth}{!}{%
    \begin{tabular}{p{3cm}p{2.5cm}p{2cm}cc}
        \toprule 
        Task & Method & $\#$ Params & NDS$^{\hat{*}}$$\uparrow$ & mAP$\uparrow$ \\
        
        \midrule
        \multirow{5}{*}{nuScenes $\rightarrow$ Lyft} 
        & \textit{Oracle} & 33.5M & 0.635 & 0.534 \\ 
        & \textit{Direct Transfer} & 33.5M & 0.338 & 0.245 \\
        \cmidrule{2-5}
        & Full FT & 33.5M & 0.638 & 0.533 \\ 
        \cmidrule{2-5}
        \coloredrowcell{DCDCDC} & Ours (50$\%$) & 12.2M & 0.596 & 0.477 \\
        \coloredrowcell{DCDCDC} & Ours (100$\%$) & 12.2M & 0.638 & 0.534 \\
        \bottomrule
    \end{tabular}}
\end{table}

\myparagraph{Comparison of UDGA on BEVFormer.}
To demonstrate the validation of UDGA, we further compare performance on BEVFormer-small (33.5M parameters) with Full FT.
For accurate comparison, we provide \textit{Oracle}, and \textit{Direct Transfer} in nuScenes $\rightarrow$ Lyft task.

BEVFormer adopt Query-based view transformation modules $\mathcal{V}$ as follow Eq.~\ref{eq:query}:
\begin{equation}
\label{eq:query}
    \mathcal{V}(I,K,T) = CrossAttn(q:P_{xyz}, k~v: F_{2d}),
\end{equation}
where $q$, $k$ and $v$ represents query, value and key in Transformer, and then $P_{xyz}$ denotes pre-defined anchor BEV positions by $K$, and $T$. Here, Query-based module benefits from $CrossAttn$ with sparse query sets, implicitly learning geometric information.
Thus, we reconstruct our UDGA paradigm without explicit depth constraints.
First, we adopt linear-based bottleneck structures with Layer Normalization in Eq.~\ref{eq:adapter2}. $\phi_{up}$ and $\phi_{down}$ denote the projection up and down layer.
\begin{equation}
\label{eq:adapter2}
    y = \textit{ffn}(x) + \phi_{up}(\sigma(\phi_{down}(LN(x)))),
\end{equation}
where \textit{ffn} denotes feed-forward networks, and $LN$ represents Layer Normalization. 
We conduct experiments by plugging these extra modules, which accounts for 36$\%$ of the total parameters, into BEVFormer.
As a result, we achieve significant adaptation performance with the 50$\%$ data split.
Notably, we demonstrate effectiveness, achieving parity with Full FT in the 100$\%$ data split.

\begin{figure}
    \centering
    \includegraphics[width=1\linewidth]{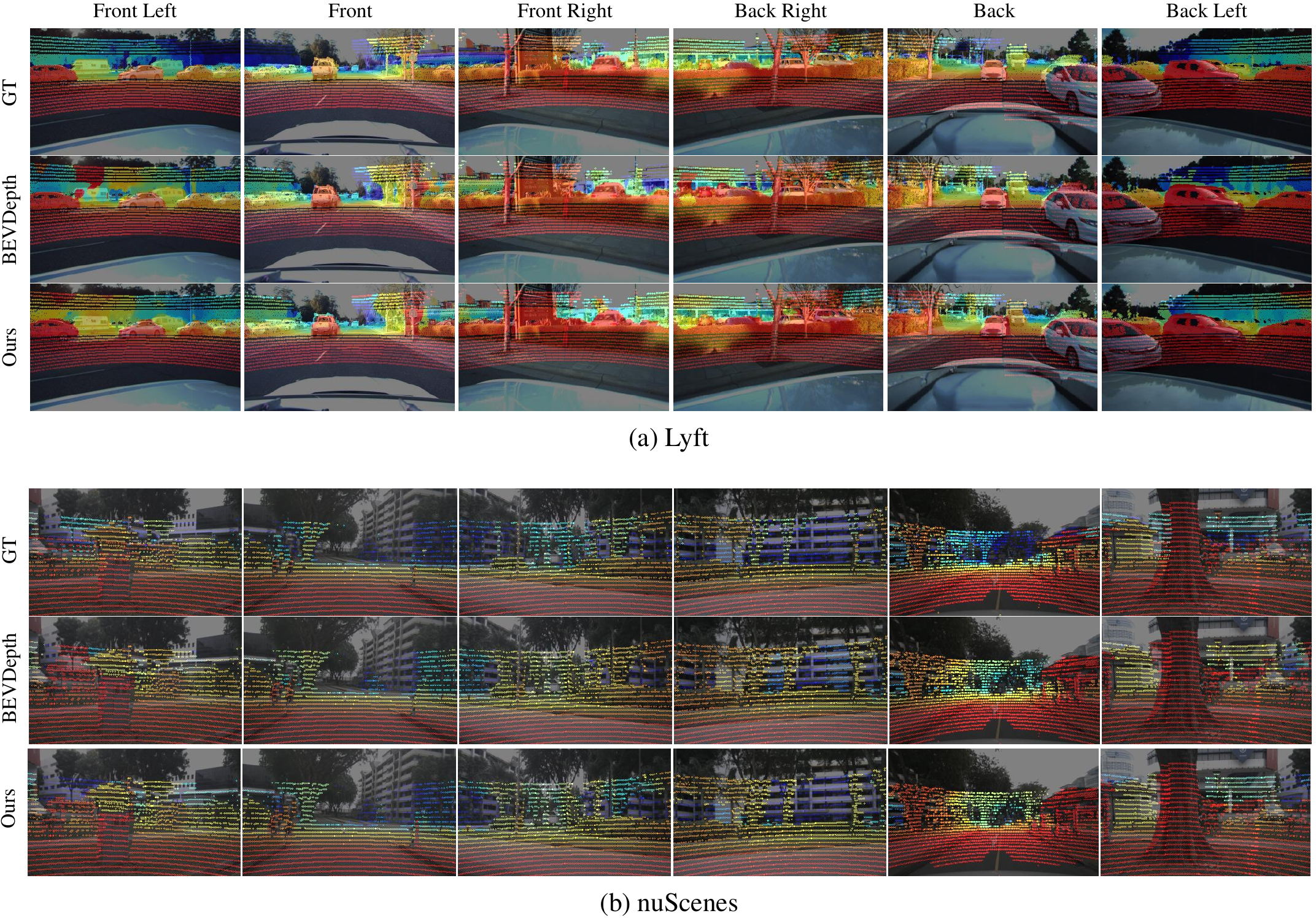}
    \caption{Multi-view visualization of the depth estimation of BEVDepth and Ours for (a)Lyft and (b)nuScenes samples. In general, our depth consistency was better in the Lyft dataset, while it was difficult to make a quantitative comparison in the case of nuScenes due to the sparseness of the LiDAR point clouds. The depth range is from 1m to 60m. Best viewed in color.}
    \label{fig:app_depth}
\end{figure}

\myparagraph{Additional qualitative analysis.}
In this section, we further visualize our depth quality in various scenarios (\ie, Lyft, and nuScenes).
Not only our overlap depth constraint significantly improve depth consistency in occluded regions, but also show better spatial understanding for hard samples (\eg, far and low distinguishable objects).
Additionally, we note that our method effectively complement insufficient contextual recognition caused by sparse depth gt in Fig.~\ref{fig:app_depth} (b).
Overall, we stably deploy Multi-View 3DOD by leveraging effective association between adjacent views.

\section{Broader Impacts.}
Our framework is a practical AI algorithm that enhances its generalization ability to handle domain changes robustly, enabling us to effectively reduce data costs and computing resources required for adaptation. Practically, our method makes it suitable for deployment in mass-produced vehicles, where the algorithm can inherit the knowledge of well-trained pretrained weights while self-learning to adapt to each fleet environment. The adaptation learning process is also simplified, making it easier to transfer improved pretrained networks. Furthermore, by demonstrating superior performance compared to previous methods that relied on LiDAR for auxiliary depth networks, our approach reduces the dependency on lidar modality. This suggests the feasibility of excluding expensive LiDAR sensors from future autonomous vehicles. 

\end{appendices}


\end{document}